%% file: paper_CL.tex
\documentclass[fullname]{clv2} 

\usepackage{amsmath,amssymb}
\usepackage{booktabs}
\usepackage{caption}
\usepackage{color}
\usepackage{graphicx}
\usepackage{pbox}
\usepackage{subcaption}

\makeatletter
\newcommand{\@BIBLABEL}{\@emptybiblabel}
\newcommand{\@emptybiblabel}[1]{}
\makeatother

\usepackage[hidelinks]{hyperref}

\usepackage{url}
\newcommand{\n}[0]{n}

\definecolor{burntorange}{rgb}{0.8, 0.33, 0.0}

\newcommand{\hsic}[0]{\text{HSIC}}
\newcommand{\mi}[0]{Moran's~I}
\newcommand{\mt}{Mantel test}

\newcommand{\etal}[0]{\textit{et al.}}
\newcommand{\ex}[1]{\textit{#1}}
\renewcommand{\vec}[1]{\mathbf{#1}}
\newcommand{\trans}[1]{#1^{\top}}
\newcommand{\specialcell}[2][c]{%
  \begin{tabular}[#1]{@{}l@{}}#2\end{tabular}}
\usepackage[english]{babel}
\usepackage[captionskip=3pt]{floatrow}
\newfloatcommand{capbtabbox}{table}[\captop][\FBwidth]  

\newcommand{\new}{}

\issue{X}{X}{2016}

\newcommand\mynewpage{}

\runningtitle{A Kernel Independence Test for Geographical Language Variation}

\runningauthor{Nguyen and Eisenstein}

\begin{document}

\title{A Kernel Independence Test for Geographical Language Variation}

\author{Dong Nguyen}
\affil{University of Twente}

\author{Jacob Eisenstein}
\affil{Georgia Institute of Technology}

\maketitle

\input{abstract-cl}
\input{intro}

\input{methods}

\input{synth}
\input{empirical}
\input{conclusion}

\section*{Acknowledgments}
Thanks to Jack Grieve for sharing the corpus of dialect variables from Letters to the Editor in North American newspapers, Arthur Gretton for advice about how best to use HSIC, Erik Tjong Kim Sang for help on using the SAND data, the DB group of the University of Twente for sharing the Dutch geotagged tweets, and Leonie Cornips and Sjef Barbiers for advice on selecting the Dutch linguistic variables.
The first author was supported by the Netherlands Organization for Scientific Research (NWO), grant 640.005.002 (FACT). The second author was supported by National Science Foundation award RI-1452443, and by the Air Force Office of Scientific Research.

\starttwocolumn
\bibliographystyle{fullname}
\bibliography{cites}

\end{document}

%% file: abstract-cl.tex
\begin{abstract}
Quantifying the degree of spatial dependence for linguistic variables is a key task for analyzing dialectal variation. However, existing approaches have important drawbacks. \new{First, they are based on parametric models of dependence, which limits their power in cases where the underlying parametric assumptions are violated.} Second, they are not applicable to all types of linguistic data: some approaches apply only to frequencies, others to boolean indicators of whether a linguistic variable is present. 
We present a new method for measuring geographical language variation, which solves both of these problems. Our approach builds on Reproducing Kernel Hilbert space (RKHS) representations for nonparametric statistics, and takes the form of a test statistic that is computed from pairs of individual geotagged observations without aggregation into predefined geographical bins. We compare this test with prior work using synthetic data as well as a diverse set of real datasets: a corpus of Dutch tweets, a Dutch syntactic atlas, and a dataset of letters to the editor in North American newspapers. Our proposed test is shown to support robust inferences across a broad range of scenarios and types of data.
\end{abstract}

%% file: intro.tex
\section{Introduction}

\begin{figure}
\centering
\includegraphics[width=2.8in]{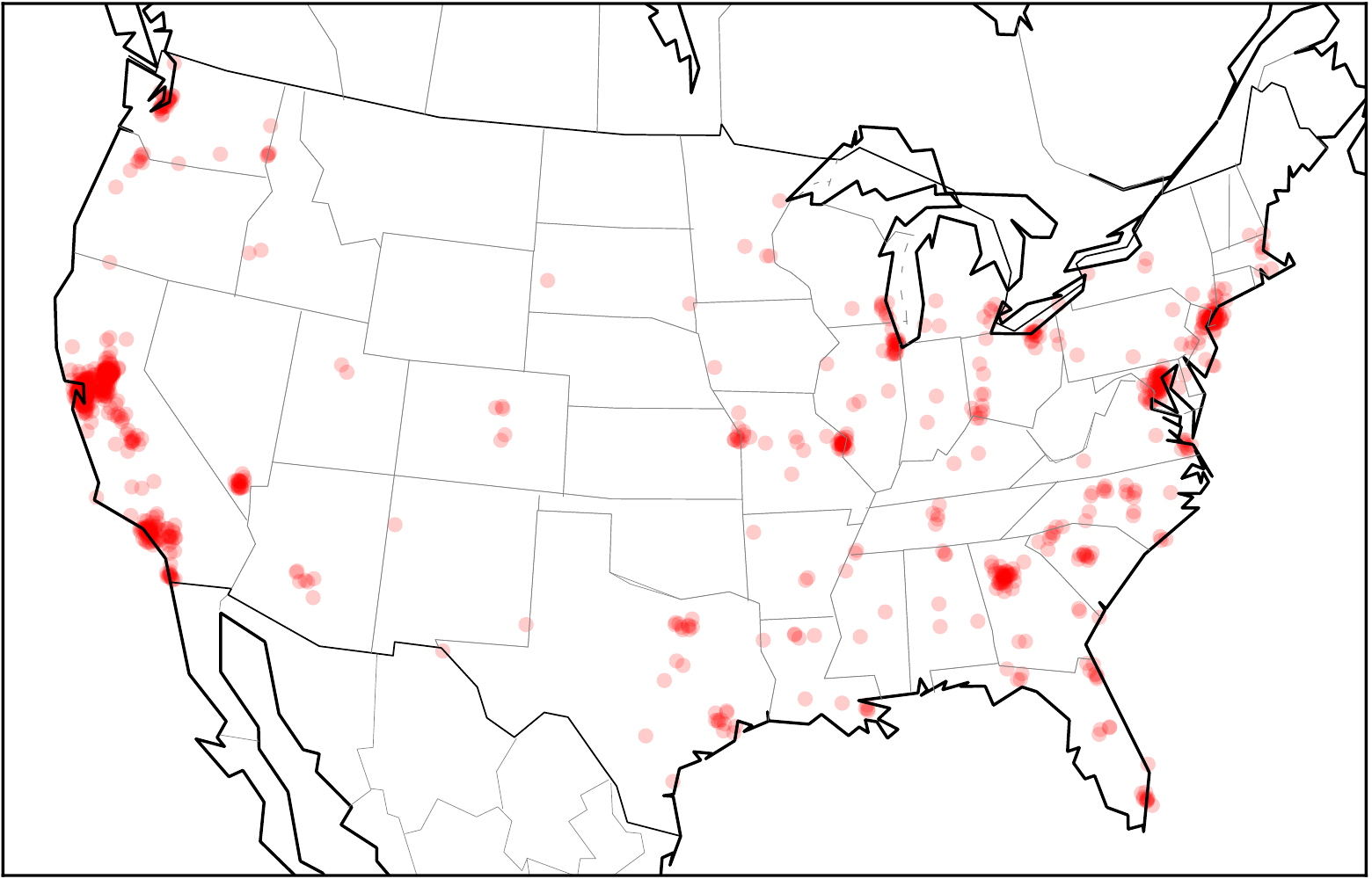}
\caption{1000 geolocated tweets containing the word \ex{hella}}
\label{fig:hella}
\end{figure}

\autoref{fig:hella} shows the geographical location of 1000 Twitter posts containing the word \ex{hella}, an intensifier used in expressions like \ex{I got hella studying to do} and \ex{my eyes got hella big}~\cite{eisenstein2014diffusion}. While the word appears in major population centers throughout the United States, the map suggests that it enjoys a particularly high level of popularity on the west coast, in the area around San Francisco. But does this represent a real geographical difference in American English, or is it the result of chance fluctuation in a finite dataset?

Regional variation of language has been extensively studied in sociolinguistics and dialectology~\cite{chambers1998dialectology,grieve2011statistical,JLG:8992970,lee1993spatial,nerbonne2013dialectometry,szmrecsanyi2012grammatical}. A common approach involves mapping the geographic distribution of a linguistic variable (e.g., the choice of \ex{soda}, \ex{pop}, or \ex{coke} to refer to a soft drink) and identifying boundaries between regions based on the data. The identification of linguistic variables that exhibit regional variation is therefore the first step in many studies of regional dialects. Traditionally, this step has been based on the manual judgment of the researcher; depending on the quality of the researcher's intuitions, the most interesting or important variables might be missed. 

The increasing amount of data available to study dialectal variation suggests a turn towards data-driven alternatives for variable selection. For example, researchers can mine social media data such as Twitter~\cite{doyle2014mapping,eisenstein2010latent,huang2015understanding} or product reviews~\cite{Hovy2015} to identify and test thousands of dialectal variables. Despite the large scale of available data, the well-known ``long tail'' phenomenon of language ensures that there will be many potential variables with low counts. A statistical metric for comparing the strength of geographical associations across potential linguistic variables would allow linguists to determine whether finite geographical samples --- such as the one shown in \autoref{fig:hella} --- reveal a statistically meaningful association.

The use of statistical methods to analyze spatial dependence has been only lightly studied in sociolinguistics and dialectology. Existing approaches employ classical statistics such as Moran's I~ (e.g., \namecite{grieve2011statistical}), join count analysis (e.g., \namecite{lee1993spatial}) and the Mantel Test (e.g., \namecite{scherrer:2012:LINGVIS2012}); we review these statistics in \autoref{sec:prior}. \new{These classical approaches suffer from a common problem: each type of test can capture only a specific parametric form of spatial linguistic variation. As a result, these tests can incorrectly fail to reject the null hypothesis if the nature of the geo-linguistic dependence does not match the underlying assumptions of the test.} 

To address these limitations, we propose a new test statistic that builds on a rich and growing literature on \new{kernel embeddings} for non-parametric statistics~\cite{shawe2004kernel}. In these methods, probability distributions, such as the distribution over geographical locations for each linguistic variable, are embedded in a Reproducing Kernel Hilbert space (RKHS). Specifically, we employ the Hilbert-Schmidt Independence Criterion (\hsic; \namecite{gretton2005measuring}). \new{Due to its ability to compare arbitrarily high-order moments of probability distributions, \hsic\ can be used to compare arbitrary probability measures, by computing kernel functions on finite samples. Unlike prior approaches, \hsic\ is statistically consistent: in the limit of a sufficient amount of data, it will correctly determine whether the distribution of a linguistic feature is geographically dependent. As a further convenience, because it is built on kernel similarity functions, \hsic\ can be applied with equal ease to any type of linguistic data, as long as an appropriate kernel function can be constructed.}

To validate this approach, we compare it against three alternative spatial statistics: Moran's I, the Mantel test, and \new{join count analysis}. For a controlled comparison, we use synthetic data to simulate different types of regional variation, and different types
of linguistic data. This allows us to measure the capability of each approach to recover true geo-linguistic associations, and to avoid Type I errors even in noisy and sparse data. Next, we apply these approaches to three real linguistic datasets: a corpus of Dutch tweets, a Dutch syntactic atlas and letters to the editor in North American newspapers.

To summarize, the contributions of this article are:
\begin{itemize}
\item We show how the Hilbert-Schmidt Independence Criterion can be applied to linguistic data. HSIC is a non-parametric test statistic, which can handle both frequency and categorical data, requires no discretization of geographic data, and \new{is capable of detecting arbitrary geo-linguistic dependencies} (\autoref{sec:hsic}).
\item We use synthetic data to compare the power and calibration of HSIC against three alternatives: \mi, the Mantel Test, and \new{join count analysis} (\autoref{sec:synth}).
 \item We apply these methods to analyze dialectal variation in three empirical datasets, in both English and Dutch, across a variety of registers (\autoref{sec:real}). 
\end{itemize}

%% file: methods.tex
\section{Prior Work}
\label{sec:prior}
This section describes prior work on \new{global} methods for quantifying the degree of spatial dependence in a geotagged corpus.\footnote{\new{Global methods test for dependence over the entire dataset. In some cases, there will be local dependence in a few ``hot spots'', even when global dependence is not detected, and local autocorrelation statistics have been proposed to capture such dependences~\cite{anselin1995local}. For example, \namecite{grieve2016regional} uses the Getis-Ord $G_i$ statistic~\cite{getis1992analysis} in his analysis of regional American English. Local statistics are particularly useful as an exploratory tool, but Grieve argues that the associated $p$-values are difficult to interpret due to the issue of multiple comparisons. We therefore focus on global tests in this paper. The adaptation of the proposed \hsic\ statistic into a local measure of dependence is an intriguing topic for future work.}} \new{While other global spatial statistics exist, we focus on the following three methods because they have been} used in previous work on dialect analysis: \mi~\cite{grieve2011statistical},  \new{join count analysis}~\cite{lee1993spatial} and the Mantel test~\cite{scherrer:2012:LINGVIS2012}. 

We define a consistent notation across methods. Let $x_i$ represent a scalar linguistic observation for unit $i \in \{1\ldots \n\}$ (typically, the presence or frequency of a linguistic variable), and let $y_i$ represent a corresponding geolocation. For convenience, we define $d_{ij}$ as the spatial distance between $y_i$ and $y_j$. Suppose we have $n$ observations, so that the data $\mathcal{D} = \{(x_1,y_1),(x_2,y_2), \ldots, (x_\n,y_\n)\}$. Our goal is to test the strength of association between $X$ and $Y$, against the null hypothesis that there is no association. 

\mynewpage
\subsection{Moran's I}
\namecite{grieve2011statistical} introduced the use of \mi~\cite{cliff1981spatial,moran1950} in the study of dialectal variation. To define the statistic, let $W = \{w_{ij}\}_{i,j \in \{1 \ldots \n\}}$ represent a \emph{spatial \new{weighting} matrix}, such that larger values of $w_{ij}$ indicate greater proximity, and $w_{ii} = 0$. In their application of \mi\ to a corpus of \new{newspaper} letters-to-the-editor, Grieve \etal\ define $W$ as,
\begin{equation}
w_{ij} = \begin{cases}
1, & d_{ij} < \tau, i\neq j \\
0, & d_{ij} \geq \tau, \text{ or } i = j
\end{cases}
\end{equation}
where $\tau$ is some critical threshold~\cite{grieve2011statistical}. When the spatial weighting matrix is defined in this way, \mi\ can be seen as a statistic that quantifies whether observations $x_i$ and $x_j$ are more similar when $w_{ij} = 1$ than when $w_{ij} = 0$.\footnote{\new{The matrix $W$ can be defined in other ways. We can define a continuous-valued version of $W$ by setting $w_{ij} = \exp(- \gamma d_{ij})$, with $d_{ij}$ equal to the geographical distance between units $i$ and $j$. Alternatively, we could define a topological spatial weighting matrix by setting $w_{ij} = 1$ when $j$ is one of the $k$ nearest neighbors of $i$~\cite{getis2010constructing}.}}

\mi\ is based on a hypothesized autoregressive process $X = \rho W X + \epsilon$, where $X$ is a vector of the linguistic observations $x_1, \ldots x_\n$, and $\epsilon$ is a vector of uncorrelated noise. Since $X$ and $W$ are given, the estimation problem is to find $\rho$ so as to minimize the magnitude of $\epsilon$. To take a probabilistic interpretation, it is typical to assume that $\epsilon$ consists of independent and identically-distributed (IID) normal random variables with zero mean~\cite{ord1975estimation}. \new{Under the null hypothesis of no spatial dependence between the observations in $X$, we would have $\rho = 0$. Note, however, that we may fail to reject the possibility that $\rho=0$ even in the presence of spatial dependence, if the form of this dependence is not monotonic or nonlinear in $W$.}

Because $\rho$ is difficult to estimate exactly~\cite{ord1975estimation}, \mi\ is used as an approximation. It is computed as,
\begin{align}
I = & \frac{\n}{\sum_i^\n (x_i - \overline{x})^2}
\frac{\sum_i^\n \sum_j^\n w_{ij} (x_i - \overline{x})
(x_j - \overline{x})}{\sum_i^\n \sum_j^\n w_{ij}},
\label{eq:mi}
\end{align}
where $\overline{x} = \frac{1}{\n}\sum_i x_i$. The ratio on the left is the inverse of the variance of $X$; the ratio on the right  corresponds to the covariance between points $i$ and $j$ that are spatially similar. Thus, the statistic rescales a spatially-reweighted covariance (the ratio on the right of \autoref{eq:mi}) by the overall variance (the ratio on the left of \autoref{eq:mi}), giving an estimate of the overall spatial dependence of $X$. A compact alternative notation is to rewrite the statistic in terms of the vector of \emph{residuals} $R = \{r_i\}_{i\in 1\ldots \n}$, where the residual $r_i = x_i - \overline{x}$. This yields the form $I = \frac{\trans{R} W R}{\trans{R} R}$, with $\trans{R}$ indicating the transpose of the column vector $R$, \new{and with $W$ assumed to be normalized so that $\sum_{i,j}w_{ij} = n$}.
\mi\ values often lie between $-1$ and $1$, but the exact range depends on the weight matrix $W$, and is theoretically unbounded~\cite{GEAN:GEAN797}. 

In hypothesis testing, our goal is to determine the $p$-value representing the likelihood that a value of \mi\ at least as extreme as the observed value would arise by chance under the null hypothesis. 
The expected value of \mi\ in the case of no spatial dependence is $-\frac{1}{n-1}$. Grieve \etal~compute $p$-values from a closed-form approximation of the variance under the null hypothesis of total randomization. A non-parametric alternative is to perform a \emph{permutation test}, calculating the empirical $p$-value by comparing the observed test statistic against the values that arise across multiple random permutations of the original data.

\new{In either case, \mi\ does not test the null hypothesis of no statistical dependence between the linguistic features $X$ and the geo-coordinates $Y$. Rather, it tests whether the estimated value of $\rho$ would be likely to arise if there were no such dependence. But if the nature of the geo-linguistic dependence defies the assumptions of the statistic, then we risk incorrectly failing to reject the null hypothesis, a type II error. Put another way, there are forms of strong spatial dependence for which $\rho=0$, such as non-monotonic spatial relationships. This risk can be somewhat ameliorated by careful choice of the spatial weighting matrix $W$, which could in theory account for non-linear or even non-monotonic dependencies. However an exhaustive search for some $W$ that obtains a low $p$-value would clearly be an invalid hypothesis test, and so $W$ must be fixed \emph{before} any test is performed. In some cases, the researcher may bring substantive insights to the determination of $W$, and so the flexibility of \mi\ in this sense could be regarded as a positive feature. But there is little theoretical guidance, and a poor selection of $W$ will result in inflated type II error rates.}

\new{From a practical standpoint, \mi\ is applicable to only some types of linguistic data.} In the study of dialect, $X$ typically represents the frequency or presence of some linguistic variable, such as the use of \ex{soda} versus \ex{pop}. We are unaware of applications of Moran's I to variables with more than two possibilities (e.g., \ex{soda}, \ex{pop}, \ex{coke}). \new{One possible solution, proposed by one of the reviewers, would be to perform multiple tests, with each alternant pitted against all the others. But it is not clear how the $p$-values from these multiple tests should be combined. For example, selecting the minimum $p$-value across the alternants would mean that the null hypothesis would be more likely to be rejected for variables with more alternants; averaging the $p$-values across alternants would have the opposite problem.} 

\subsection{Join Count Analysis}
\new{If the linguistic data $X$ consists of discrete observations, a simple approach  is the use of \emph{join count analysis}. For each pair of points $(i,j)$, we compute $w_{ij} \delta(x_i = x_j)$, where $\delta(x_i = x_j)$ returns a value of $1$ if $x_i$ and $x_j$ are identical, and $0$ otherwise. As in \mi, $w_{ij}$ is an element of a spatial weighting matrix, which could be binary or continuous. The global sum of the counts is computed as,
\begin{align}
\text{num-agree} = & \sum^\n_i \sum^\n_j w_{ij}(x_i x_j + (1-x_i)(1-x_j))\\
= & \trans{X} W X + \trans{(1-X)} W (1-X),
\end{align}
with $\trans{X}$ indicating the transpose of the column vector $X$. Note the similarity to the numerator of \mi, which can be written as $\trans{R} W R$. The number of agreements can be compared with its expectation under the null hypothesis, yielding a hypothesis test for global autocorrelation~\cite{cliff1981spatial}.}

\new{Join count analysis has been applied to the study of dialect by~\namecite{lee1993spatial}, who take each linguistic observation $x_i \in \{0,1\}$ to be a binary variable indicating the presence or absence of a dialect feature. They then build a binary spatial weighting matrix by performing a Delaunay triangulation over the geolocations of participants in dialect interviews, with $w_{ij}=1$ if the edge $(i,j)$ appears in the Delaunay triangulation.} A nice property of Delaunay triangulation is that points tend to be connected to their closest neighbors, regardless of how distant or near those neighbors are: in high-density regions, the edges will tend to be short, while in low-density regions, the edges will be long. The method is therefore arguably more suitable to data in which the density of observations is highly variable --- for example, between densely-populated cities and sparse-populated hinterlands.

Because \new{join count statistics} are based on counts of agreements, this form of analysis requires that each $x_i$ is a categorical variable --- possibly non-binary --- rather than a frequency. In this sense, it is the \new{complement} of \mi, which can be applied to frequencies, but not to non-binary variables. Thus, \new{join count analysis} is best suited to cases where observations correspond to individual utterances (e.g., Twitter data, dialect interviews), rather than cases where observations correspond to longer texts (e.g., newspaper corpora). 

\subsection{The Mantel Test}
The \mt\ can in principle be used to measure the dependence between any two arbitrary signals. In this test, we compute \emph{distances} for each pair of linguistic variables, $d_x(x_i,x_j)$, and each pair of spatial locations, $d_y(y_i,y_j)$, forming a pair of distance matrices $D_x$ and $D_y$. We then estimate the element-wise correlation (usually, the Pearson correlation) between these two matrices.  \namecite{scherrer:2012:LINGVIS2012} uses the \mt\ to correlate linguistic distance with geographical distance, and~\namecite{Gooskens01112006} correlate perceptual distance with linguistic distance. The \mt\ has also been applied to non-human dialect analysis, revealing regional differences in the call structures of Amazonian parrots by computing a linguistic distance matrix $D_x$ directly from spectral measurements~\cite{wright1996regional}.

\new{Because it is built around distance functions, the \mt\ is applicable to binary, categorical, and frequency data --- any kind of data for which a distance function can be constructed.} For spatial locations, a typical choice is to compute the distance matrix based on the Euclidean distance between each pair of points. For binary or categorical linguistic data, the entries of the linguistic distances matrix can be set to $0$ if $x_i = x_j$, and $1$ otherwise. For linguistic frequency data, we use the absolute difference between the frequency values.

The role of hypothesis testing in the \mt\ is to determine the likelihood that the observed test statistic --- in this case, the correlation between the distance matrices $D_x$ and $D_y$ --- could have arisen by chance under the null hypothesis. In the ideal case of perfect correlation, twice as much geographical distance should imply twice as much as linguistic distance. \new{But this situation is highly unlikely to obtain in practice. In fact, as one of the reviewers noted, such a perfect correlation \emph{cannot} arise from any linguistic data involving a single variable: even if a linguistic variable obeys a perfect dialect continuum (e.g., varying in frequency from east to west), the distances in the orthogonal north-south direction would diminish the resulting correlation. In realistic settings in which the geo-linguistic dependence is obscured by noise, this can dramatically diminish the power of the test. Note that even non-linear transformations of the distance metric would not correct this issue. The key problem, as identified by \namecite{legendre2015should}, is that the \mt\ is not designed to test for independence between $X$ and $Y$, but rather, the correlation between \emph{distances} on $X$ and $Y$. When distances are the primary units of analysis --- as, for example, in the work of  \namecite{Gooskens01112006} --- the test is applicable. But for the task of determining whether a specific linguistic variable is geographically dependent, the test is incorrectly applied; as we show in \autoref{sec:synth}, this results in inflated Type II error rates.}

\subsection{Other Related Work}
\new{
Several computational studies attempt to characterize linguistic differences across geographical regions, although in general these studies do not perform hypothesis testing on geographical dependence. In general, these studies rely on aggregating geotagged social media content into geographical bins. Some studies rely on politically defined units such as nations and states~\cite{Hovy2015}; however, \emph{isoglosses} (the geographical boundaries between linguistic features) need not align with politically-defined geographical units~\cite{nerbonne2013dialectometry}. Other approaches rely on automatically defined geographical units, induced by computational methods such as geodesic grids~\cite{wing2011simple}, KD-trees~\cite{roller2012supervised}, Gaussians~\cite{eisenstein2010latent}, and mixtures of Gaussians~\cite{hong2012discovering}. While these approaches offer insight about the nature of geographical language variation, they do not provide test statistics that allow us to quantify the geographical dependence of various linguistic features.}

\new{As described in the next section, our approach is based on reproducing kernel Hilbert spaces, which enable us to non-parametrically compare probability distributions. Another way in which kernel methods can be applied to spatial analysis is in Gaussian Processes, which are often used to represent spatial data~\cite{cressie1988spatial,ecker1997bayesian}. Specifically, we can define a kernel over space, so that a response variable is distributed as a Gaussian with covariance defined by the kernel function. For example, it might be possible to model the popularity of linguistic features {as a Gaussian Process}, using the spatial covariance kernel to make smooth predictions at unknown locations. Our approach in this paper is different, as we are interested in hypothesis testing, rather than modeling and prediction. Another difference is that we apply kernels to both the geographical and linguistic data sources, while a Gaussian Process approach would make the parametric assumption that the linguistic signal is Gaussian distributed with covariance defined by the spatial covariance kernel.}

\mynewpage
\section{Hilbert-Schmidt Independence Criterion (HSIC)}
\label{sec:hsic}
\new{\mi, join count analysis, and the \mt\ share an important drawback: they do not directly test the independence of language and geography, but rather, they test for autocorrelation between $X$ and $Y$, or between distances on these variables. \mi\ tests whether the parameter of a linear autoregressive model is nonzero; the \mt\ is performed on the correlations between pairwise distances; join count statistics enable tests of whether spatially adjacent units tend to have the same linguistic features. In each case, rejection of the null hypothesis implies dependence between the geographical and linguistic signals. However, each test can incorrectly fail to reject the null hypothesis if its assumptions are violated, even if given an arbitrarily large amount of data.}

\new{We propose an alternative approach: directly test for the
independence of geographical and linguistic variables,
$P_{XY} \stackrel{?}{=} P_X P_Y$. Our approach, which is based on the
Hilbert-Schmidt Independence Criterion (HSIC)~\cite{gretton2005measuring,gretton2008kernel}, makes no parametric assumptions about the form of these distributions. The proposed test is \emph{consistent}, in the sense that it will always reach the right decision, if provided enough data~\cite{fukumizu2008kernel}.}\footnote{HSIC was introduced as a test of independence by \namecite{gretton2008kernel}. In this paper, we present the first application to computational linguistics.}

\new{To test independence for arbitrary distributions $P_{XY}$, $P_X$, and $P_Y$, \hsic\ employs the framework of Reproducing Kernel Hilbert spaces (RKHS). This framework will be familiar to the computational linguistics community through its application to support vector machines~\cite{collins2001convolution,lodhi2002text}, where \emph{kernel similarity functions} between pairs of instances are used to induce classifiers with highly non-linear decision boundaries. \hsic\ is a kernelized independence test, and it offers an analogous advantage: by computing kernel similarity functions on pairs of observations, it is possible to implicitly compare probability distributions across high-order moments, enabling non-parametric independence tests that are statistically consistent. An additional advantage of the RKHS framework is that it can be applied to arbitrary linguistic data --- including dichotomous, polytomous, continuous, and vector-valued observations --- as long as an appropriate kernel similarity function can be defined.} 

\newcommand{\mmd}{\textsc{mmd}}
\newcommand{\E}{\mathbb{E}}
\newcommand{\X}{\mathcal{X}}
\newcommand{\Y}{\mathcal{Y}}
\newcommand{\kdot}[1]{k(\cdot,#1)}
\newcommand{\hprod}[1]{\langle #1 \rangle}
\subsection{Comparing Probability Distributions}
\new{The \emph{maximum mean discrepancy} ($\mmd$) is a non-parametric statistic that compares two arbitrary probability distributions. In the \hsic\ test, this statistic is used to compare the joint distribution $P_{XY}$
with the product of marginal distributions $P_X P_Y$. The $\mmd$ is defined as,
\begin{equation}
\mmd(P,Q) = \sup_{f} (\E_P[f(X)] - \E_Q[f(Y)]),
\end{equation}
where we take the supremum $f$ over a set of possible functions, and compute the difference in the expected values under the distributions $P$ and $Q$. Clearly, if $P=Q$, then $\mmd=0$, but the challenge is to compute $\mmd$ for arbitrary $P$ and $Q$, based only on finite samples from these distributions.}

\new{To explain how to do this, we introduce some concepts from RKHS. Let $k : \mathcal{X} \times \mathcal{X} \mapsto \mathbb{R}_+$ denote a kernel function, mapping from pairs of observations $(x_i, x_j)$ to non-negative reals. A classical example is the radial basis function (RBF) kernel on vectors, where $k(\vec{x}_i, \vec{x}_j) = \exp \left(-\gamma || \vec{x}_i - \vec{x}_j||_2^2\right)$. Many other kernel functions are possible; the conditions for valid kernels are enumerated in \autoref{sec:kernels}.}

\new{For any instance $x$, the kernel function $k$ defines a corresponding \emph{feature map} $\kdot{x} : \X \mapsto \mathbb{R}_+$, which is simply the function that arises by fixing one of the arguments of the kernel function to the value $x$. The ``reproducing'' property of RKHS implies an identity between kernel functions and inner products of feature maps:
\begin{equation}
k(x_i,x_j) = \hprod{\kdot{x_i}, \kdot{x_j}}.
\end{equation}
Thus, even though the feature map may be an arbitrarily complex function of $x$, we can compute inner products of feature maps by computing the kernel similarity function over the associated instances. The $\mmd$ can be expressed in terms of such inner products, and thus, can be computed in terms of kernel similarity functions.}

\new{For a probability measure $P$, the \emph{mean element} of $P$ is defined as the expected feature map, $\mu_P = \E_P \kdot{x}$. The $\mmd$ can then be computed in terms of kernel functions of the mean elements,
\begin{equation}
\mmd^2(P,Q) = \hprod{\mu_P, \mu_P} + \hprod{\mu_Q,\mu_Q} - 2\hprod{\mu_P,\mu_Q}.
\label{eq:mmd}
\end{equation}}

\new{If $\mu_P = \mu_Q$, then the $\mmd$ is zero. The key observation is that $\mu_P = \mu_Q$ if and only if $P=Q$, so long as an appropriate kernel similarity function is chosen~\cite{fukumizu2008kernel}; see \autoref{sec:kernels} for more on the choice of kernel functions.}

\new{Each of the inner products in \autoref{eq:mmd} corresponds to an expectation that can be estimated empirically from finite samples $\{x_1, x_2, \ldots, x_m\}$ and $\{y_1, y_2, \ldots, y_\n\}$,
\begin{align}
\mmd^2(P,Q) = & \E_{x,x' \sim P} k(x,x') + \E_{y,y'\sim Q} k(y,y') - 2 \E_{x\sim P, y\sim Q} k(x,y) \\
\widehat{\mmd^2}(P,Q)= & \frac{1}{m^2}\sum_{i,j}^m k(x_i,x_j)
+ \frac{1}{n^2}\sum_{i,j}^n k(y_i,y_n)
- \frac{2}{nm} \sum_{i=1}^m\sum_{j=1}^n k(x_i,y_j).
\label{eq:mmd-finite}
\end{align}
The full derivation is provided by \namecite{gretton2008kernel}. Having shown how to estimate a statistic on whether two probability measures are identical, we now use this statistic to test for independence.}

\subsection{Derivation of HSIC}
\new{To construct an independence test over random variables $X$ and $Y$, we test the $\mmd$ between the joint distribution $P_{XY}$ and the product of marginals $P_X P_Y$. In this setting, each observation $i$ corresponds to a pair $(x_i,y_i)$, so we require a kernel function on paired observations, $k((x_i,y_i),(x_j,y_j))$. We define this as a \emph{product kernel},
\begin{equation}
k((x_i,y_i),(x_j,y_j)) = k_{\X}(x_i,x_j) k_{\mathcal{Y}}(y_i,y_j),
\end{equation}
where $k_\X$ and $k_\Y$ are kernels for the linguistic and geographic observations respectively.}

\new{Using the product kernel, we can define mean embeddings for the distributions $P_{XY}$ and $P_X P_Y$, enabling the application of the $\mmd$ estimator from \autoref{eq:mmd-finite}. The Hilbert-Schmidt Independence Criterion (\hsic) is precisely this application of maximum mean discrepancy to compare the joint distribution against the product of marginal distributions.}

\new{Concretely, let us define the \emph{Gram matrix} $K_x$ so that $(K_x)_{i,j} = k_\X(x_i,x_j)$ for all pairs $i,j$ in the sample. Analogously, $(K_y)_{i,j} = k_{\mathcal{Y}}(y_i,y_j)$. Then the \hsic\ can be estimated from a finite sample of $m$ observations as 
\begin{align}
\widehat{\hsic} = & \frac{1}{\n^2} \sum_{i,j}^m (K_x)_{i,j} (K_y)_{i,j} +\frac{1}{\n^4}\sum_{i,j,q,r}^m (K_x)_{i,j} (K_y)_{q,r} -\frac{2}{\n^3}\sum_{i,j,q}^m (K_x)_{i,j} (K_y)_{i,q}\\
= & \frac{\mathbf{tr} K_\X H K_\Y H}{\n^2},
\end{align}
where $\mathbf{tr}$ indicates the matrix trace and $H$ is a centering matrix, $H = \mathbb{I}_m - \frac{1}{\n}\mathbf{1}\mathbf{1}^{\top}$. With this definition of $H$, we have,
\begin{align}
 (K_{\X}H)_{ij} = & k_\X(x_i,x_j) -  \frac{1}{\n}\sum_{j'}k_{\X}(x_i,x_{j'})\\
 (K_{\Y}H)_{ij} = & k_\Y(y_i,y_j) - \frac{1}{\n}\sum_{j'}k_{\Y}(y_i,y_{j'}).
\end{align}}

These two terms can be seen as mean-centered Gram matrices. By computing the trace of their matrix product, we obtain a cross-covariance between the Gram matrices. \new{This trace is directly proportional to the maximum mean discrepancy between $P_{XY}$ and $P_X P_Y$. If $X$ and $Y$ are dependent, then large values of $k_\X(x_i,x_j)$ will imply large values of $k_\Y(y_i,y_j)$ --- similar geography implies similar language --- and so the cross-covariance will be greater than zero. If $X$ and $Y$ are independent, then large values of $k_\X(x_i,x_j)$ are not any more likely to correspond to large values of $k_\Y(y_i,y_j)$, and so the expectation of this cross-covariance will be zero.}

\subsection{Kernel Functions}
\label{sec:kernels}
\new{To apply \hsic\ to the problem of detecting geo-linguistic dependence, we must define the kernel functions $k_\X$ and $k_\Y$.
In the RKHS framework, valid kernel functions must be symmetric and positive semi-definite. To ensure consistency of the kernel-based estimator for $\mmd$, the kernel must also be \emph{characteristic}, meaning that it induces an injective mapping between probability measures and their corresponding mean elements~\cite{fukumizu2008kernel}: each probability measure $P$ must correspond to a single unique mean element $\mu_P$. Muandet \etal\ elaborate these and other properties of several well known kernels~\cite[Table 3.1]{muandet2016kernel}.}

\new{For the spatial kernel $k_{\Y}$, we employ a Gaussian radial basis function (RBF), which is a widely used choice for vector data. Specifically, we define $k_\Y(y_i,y_j; \gamma) = \exp(-\gamma d^2_{i,j})$, where $d^2_{ij}$ is the squared Euclidean distance between $y_i$ and $y_j$, and $\gamma$ is a parameter of the kernel function. We also employ the RBF in $k_\X$ when the linguistic observations take on continuous values, such as frequencies or phonetic data. The RBF kernel is symmetric, positive semi-definite, and characteristic.}\footnote{Flaxman recently proposed a ``kernelized Mantel test'', in which correlations are taken between kernel similarities rather than distances~\cite{flaxman2015machine}. The resulting test statistic is similar, but not identical to HSIC. Specifically, while HSIC centers the kernel matrix against the local mean kernel similarities for each point, the kernelized Mantel test centers against the global mean kernel similarity. This makes the test more sensitive to distant outliers. We implemented the kernelized Mantel test, and found its performance to be similar to the classical Mantel test, with lower statistical power than HSIC. Flaxman made similar observations in his analysis of the spatiotemporal distribution of crime events.}

\new{The parameter $\gamma$ corresponds to the ``length-scale'' of the kernel. Intuitively, as this parameter increases, the kernel similarity drops off more quickly with distance. In this paper, we follow the popular heuristic of setting $\gamma$ to the median of the data $(y_1, \ldots y_\n)$, as proposed by~\namecite{gretton2005kernel}. We empirically test the sensitivity of \hsic\ to this parameter in \autoref{sec:synth}. More recent work offers optimization-based approaches for setting this parameter~\cite{gretton2012optimal}, but we do not consider this possibility here.}

\new{Linguistic data is often binary or categorical. In this case, we use a Delta kernel (also sometimes called a Dirac kernel). This kernel is simply defined as $k_\X(x_i,x_j)  = 1$ if $x_i = x_j$ and 0 otherwise. The Delta kernel has been used successfully in combination with \hsic\ for high-dimensional feature selection~\cite{Song2012,yamada2014high}, and is symmetric, positive semi-definite, and characteristic for discrete data.}

\subsection{Scalability}
The size of each Gram matrix is the square of the number of observations. For large datasets, this will be too expensive to compute. Following \namecite{gretton2005measuring}, we employ a low-rank approximation to each Gram matrix, using the incomplete Cholesky decomposition~\cite{bach2002kernel}. Specifically, we approximate the symmetric matrices $K_\X$ and $K_\Y$ as low-rank products, $K_\X \approx A A^T$ and $K_\Y \approx B B^T$, where $A \in \mathbb{R}^{n \times r_A}$ and $B \in \mathbb{R}^{n \times r_B}$. The approximation quality is determined by the parameters $r_A$ and $r_B$, which are set to ensure that the  magnitudes of the residuals $K - AA^T$ and $L - BB^T$ are below a predefined threshold. HSIC may then be approximated as:
\begin{align}
\widehat{\hsic} &= \frac{\mathbf{tr} K_\X HK_\Y H}{\n^2},\\
&\approx  \frac{\mathbf{tr} (A A^T)H(B B^T)H}{\n^2},\\
&=  \frac{\mathbf{tr} (B^T (HA))(B^T(HA))^T}{\n^2}
\end{align}
where the matrix product $HA$ can be computed without explicitly forming the $\n \times \n$ matrix $H$, due to its simple structure. \new{Alternative methods for scaling the computation of \hsic\ are discussed in a recent note by \namecite{zhang2016large}.}

%% file: synth.tex
\mynewpage
\section{Synthetic Data}
\label{sec:synth}
Real linguistic datasets lack ground truth about which features are geographically distinct, making it impossible to use such data to quantitatively evaluate the proposed approaches. We therefore use synthetic data to compare the power and sensitivity of the various approaches described above. Our main goals are: (1) to calibrate the $p$-values produced by each approach in the event that the null hypothesis is true, using completely randomized data; (2) to test the power of each approach to capture spatial dependence, particularly under conditions in which the spatial dependence is obscured by noise.

\new{We compare \hsic\ with specific instantiations of the methods described in \autoref{sec:prior}, focusing on previous published applications of these methods to dialect analysis. Specifically, we consider the following methods:}

\new{\begin{description}
\item[\mi] We follow \namecite{grieve2011statistical}, using a binary spatial weighting matrix with a distance threshold $\tau$, usually set to the median of the distances between points in the dataset. This method is not applicable to categorical data.
\item[Join counts] We follow the approach of ~\namecite{lee1993spatial}, who define a binary spatial weighting matrix from a Delaunay triangulation, and then compute join counts for linked pairs of observations. This method is not applicable to frequency data. 
\item[Mantel test] We use Euclidean distance for the geographical distance matrix. For continuous linguistic data, we also use Euclidean distance; for discrete data, we use a delta function.
\end{description}}

For all approaches, a one-tailed significance test is appropriate, since in nearly all conceivable dialectological scenarios we are testing only for the possibility \new{that geographically proximate units are \emph{more} similar than they would be under the null hypothesis.} For some methods, it is possible to calculate a $p$-value from the test statistic using a  closed form estimate of the variance. However, for consistency, we employ a permutation approach to characterize the null distribution over the test statistic values. We permute the linguistic data $x$, breaking any link between geography and the language data, and then compute the test statistics for many such permutations. 

\subsection{Data Generation}
\label{sec:data_generation}
To ensure the verisimilitude of our synthetic data, we target the scenario of geo-tagged tweets in the Netherlands. For each municipality $i$, we stochastically determine the number and location of the tweets as follows:
\begin{description}
\item[Number of data points] For each municipality, the number of tweets $\n_i$ is chosen to be proportional to the population, as estimated by Statistics Netherlands (CBS). Specifically, we draw $\tilde{\n}_i \sim  \text{Poisson}(\mu_{obs} * \text{population}_i)$ and 
then set $\n_i = \tilde{\n}_i + 1$, ensuring that each municipality has at least one data point. The parameter $\mu_{obs}$ controls the frequency of the linguistic variable. For example, a common orthographic variable (e.g., ``g-deletion'') might have a high value of $\mu_{obs}$, while a rare lexical variable (e.g., \ex{soda} versus \ex{pop}) might have a much lower value. Note that $\mu_{obs}$ is shared across all municipalities.
\item[Locations] Next, for each tweet $t$, we determine the location $y_t$ by sampling without replacement from the set of real tweet locations in municipality $i$ (the dataset is described in \autoref{sec:twitter}). This ensures that the distribution of geo-locations in the synthetic data matches the real geographical distribution of tweets, rather than drawing from a parametric distribution which may not match the complexity of true geographical population distributions. Each location is represented as a latitude and longitude pair.
\end{description}
For each variable, each municipality is assigned a frequency vector $\theta_i$, indicating the relative frequency of each variable form: e.g., 70\% \ex{soda}, 30\% \ex{pop}. We discuss methods for setting $\theta_i$ below, which enable the simulation of a range of dialectal phenomena. 

We simulate both counts data and frequency data. In counts data --- such as geotagged tweets --- the data points in each instance in municipality $i$ are drawn from a binomial or multinomial distribution with parameter $\theta_i$. In frequency data, we observe only the relatively frequency of each variable form for each municipality. In this case, we draw the frequency from a Dirichlet distribution with expected value equal to $\theta_i$, drawing $\phi_t \sim \text{Dirichlet}(s \theta_i)$, where the scale parameter $s$ controls the variance within each municipality.




\begin{figure}
\centering
\begin{subfigure}{0.5\textwidth}
  \centering
  \includegraphics[width=0.5\linewidth]{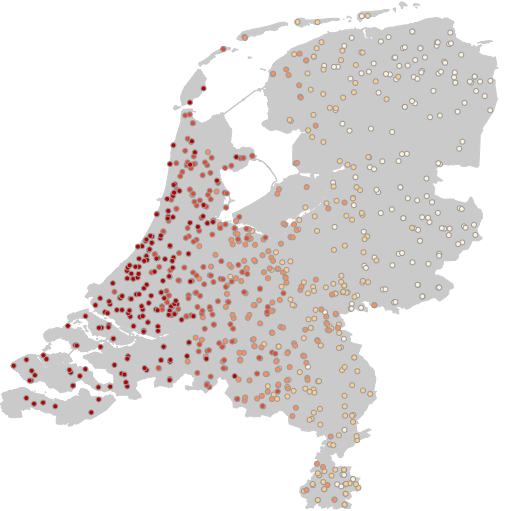}
  \caption{Angle: 0 degrees (east to west)}
\end{subfigure}%
\begin{subfigure}{.5\textwidth}
  \centering
  \includegraphics[width=0.5\linewidth]{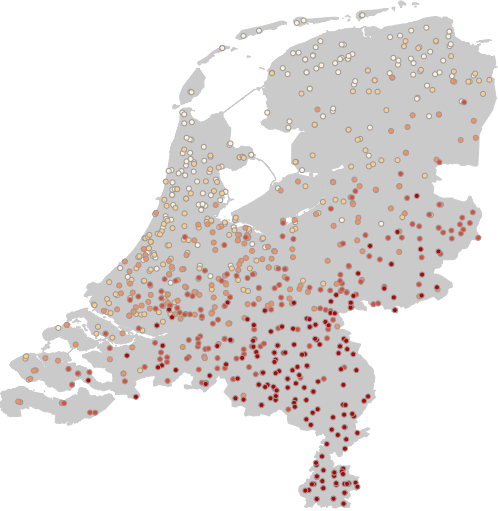}
  \caption{Angle: 120 degrees}
\end{subfigure}
\caption{Synthetic frequency data with dialect continua in two different angles} 
\label{fig:linear_angles}
\end{figure}

\mynewpage

\subsection{Calibration}
\label{sec:calibration_p_values}
Our first use of synthetic data is to examine the $p$-values obtained from each method when the null hypothesis is true --- that is, when there is no geographical variation in the data. The $p$-value corresponds to the likelihood of seeing a test statistic at least as extreme as the observed value, under the null hypothesis. Thus, if we repeatedly generate data under the null hypothesis, a well-calibrated test will return a distribution of $p$-values that is uniform in the interval $[0,1]$: for example, we expect to observe $p < .05$ in exactly 5\% of cases, corresponding to the allowed rate of Type I errors (incorrect rejection of the null hypothesis) at the threshold $\alpha = 0.05$. 



To measure the calibration of each of the proposed tests, we generate 1,000 random datasets using the procedure described above, and then compute the $p$-values under each test. In these random datasets, the relative frequency parameters $\theta_{i}$ are the same for all municipalities, which is the null hypothesis of complete randomization. To generate the binary and categorical data, we use $\mu_{obs}=10^{-5}$, meaning that the expected number of observations is one per hundred thousand individuals in the municipality or province; for comparison, this corresponds roughly to the tweet frequency of the lengthened spelling \ex{hellla} in the 2009-2012 Twitter dataset gathered by \namecite{eisenstein2014diffusion}.

\begin{figure}
\centering
\begin{subfigure}[t]{0.31\textwidth}
  \centering
  \includegraphics[width=1\linewidth]{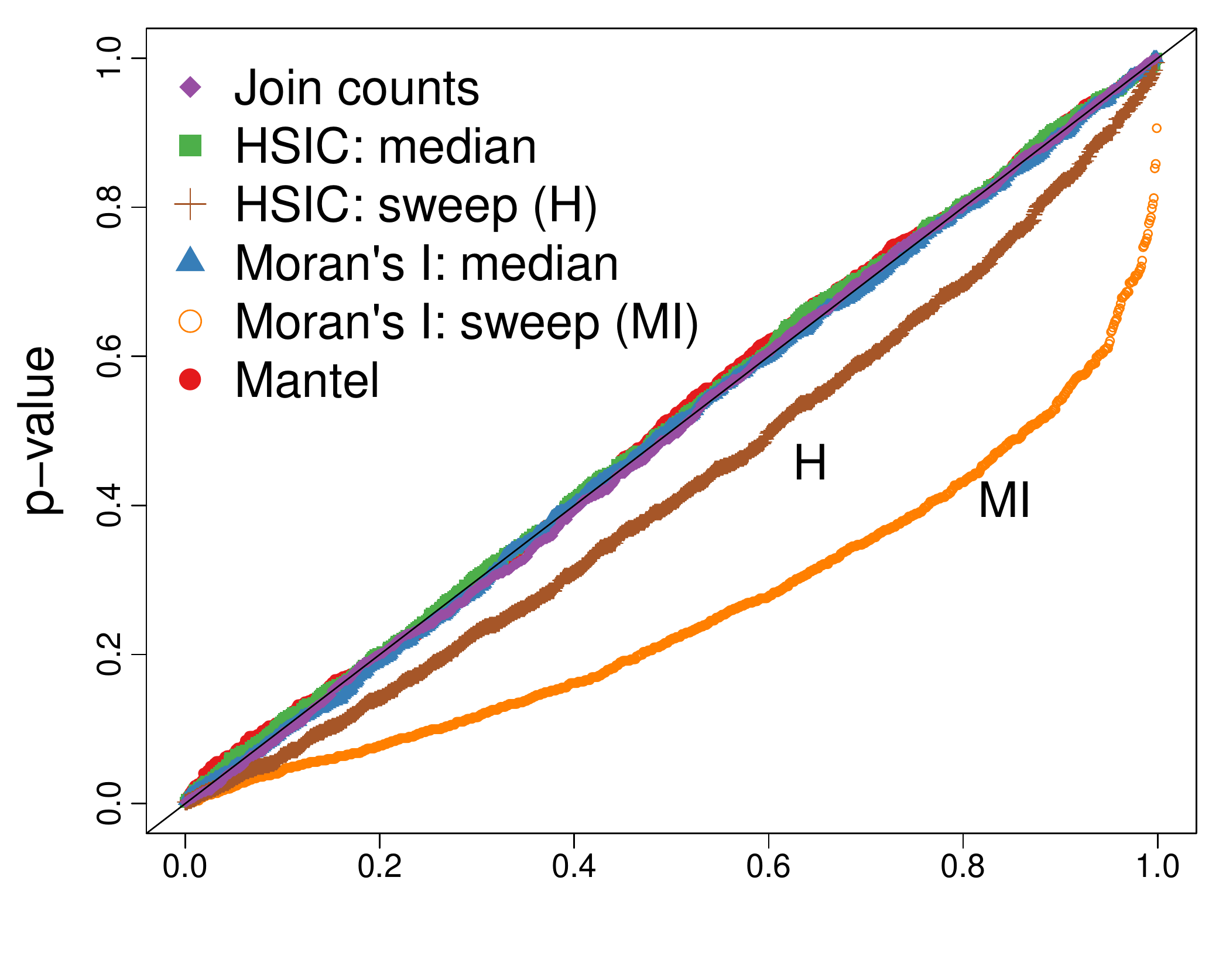}
  \caption{Binary data}
\end{subfigure}%
\hspace{0.1in}
\begin{subfigure}[t]{.31\textwidth}
  \centering
  \includegraphics[width=1\linewidth]{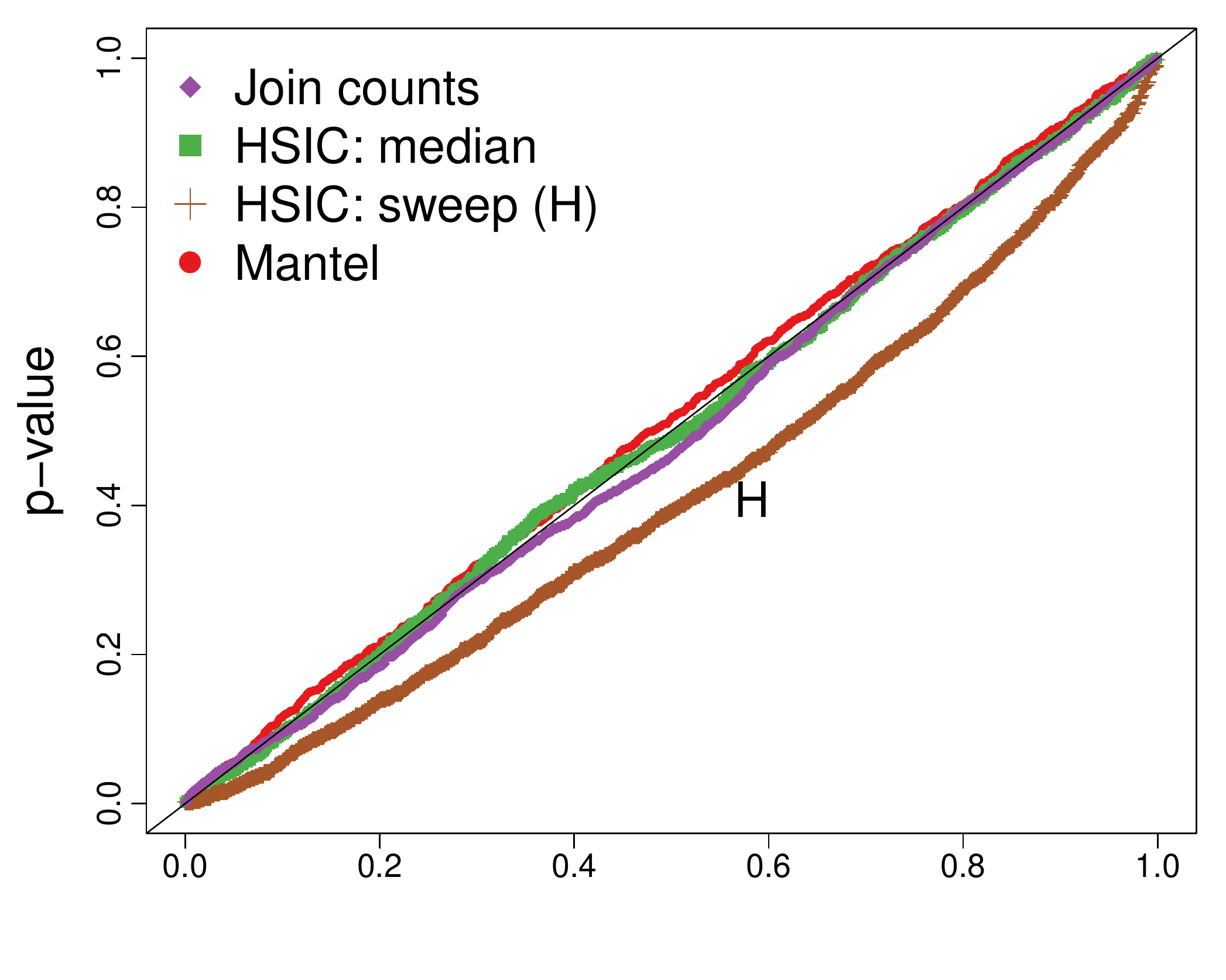}
  \caption{Categorical data}
\end{subfigure}
\hspace{0.1in}
\begin{subfigure}[t]{.31\textwidth}
  \centering
  \includegraphics[width=1\linewidth]{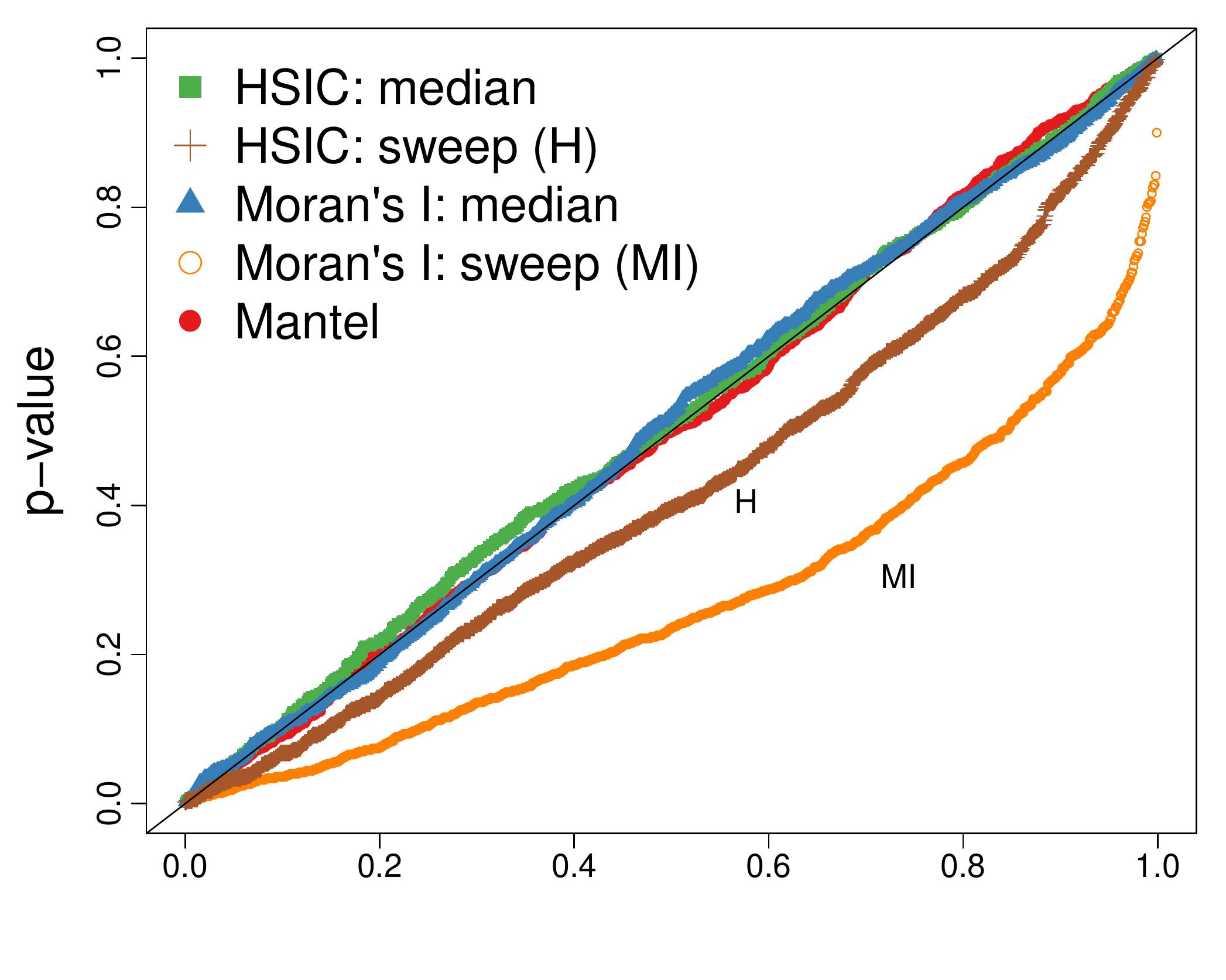}
  \caption{Frequency data\\}
\end{subfigure}
\caption{Quantile-quantile plots comparing the distribution of the obtained $p$-values with a uniform distribution. The y-axis is the $p$-value
returned by the tests. The x-axis shows the corresponding quantile
for a uniform distribution on the range [0,1]. The approaches that optimize the parameters, i.e., the cutoff for Moran's I (MI) and the bandwidth for HSIC (H), lead to a skewed distribution of $p$-values. } 
\label{fig:random_data}
\end{figure}

To visualize the calibration of each test, we use quantile-quantile (Q-Q) plots, comparing the obtained $p$-values with a uniform distribution. A well-calibrated test should give a diagonal line from the origin to $(1,1)$. \autoref{fig:random_data} shows the Q-Q plots obtained from each method on each relevant type of data (recall that not all methods can be applied to all types of data, as described in the previous section). 

\hsic~and \mi\ each have tuning parameters that control the behavior of the test: the kernel bandwidth in \hsic\, and the distance cutoff in \mi. A simple heuristic is to use the median Euclidian distance $\overline{d}$: in \mi, we use $\overline{d}$ as the distance threshold for constructing the neighborhood matrix $W$; in \hsic,
we use $\frac{1}{\overline{d}^2}$ as the kernel bandwidth parameter. \autoref{fig:random_data} shows that by basing these parameters on the median distance between pairs of points, we get well-calibrated results. However, some prior work takes an alternative approach, sweeping over parameter values to obtain the most significant results~\cite{grieve2011statistical}. In our experiments we sweep across the distance cutoff for \mi, and the bandwidth for the spatial distances in HSIC. This distorts the calibration, particularly for \mi, meaning that the resulting $p$-values are not reliable.  This is most severe for \mi\ on the municipality level, reaching type I error rates of $11.7\%$ (binary data) and $14.3\%$ (frequency data) when the significance threshold $\alpha$ is set to $5\%$. Given that such parameter sweeps are explicitly designed to maximize the number of positive test results --- and not the overall calibration of the test --- this is unsurprising. We therefore avoid parameter sweeps in the remainder of this article, and rely instead on median distance as a simple heuristic alternative.

\mynewpage
\subsection{Power}
Next, we consider synthetic data in which there is geographical variation by construction. We assess the \emph{power} of each approach by computing the fraction of simulations for which the approaches correctly rejected the null hypothesis of no spatial dependence, given a significance threshold of $\alpha=0.05$.
We again use the Netherlands as the stage for all simulations, and consider two types of geographical variation. 
\begin{description}
\item[Dialect continua] We generate data such that the frequency of a linguistic variant increases linearly through space, as in a dialect continuum~\cite{heeringa2001dialect}. In most of the synthetic data experiments below, we average across a range of angles, from $0^\circ$ to $357^\circ$ with step sizes of $3^\circ$, yielding 120 distinct angles in total. Each angle aligns differently with the population distribution of the Netherlands, so we also assess sensitivity of each method to the angle itself. Figure \ref{fig:linear_angles} shows two synthetic datasets with dialect continua in different angles.
\item[Geographical centers] Second, we consider a setting in which variation is based on one or more geographical \emph{centers}. In this setting, all cities within some specified range of the center (or one of the centers) have some maximal frequency value $\theta_i$; in other cities, this value decreases as distance from the nearest center grows. This corresponds to the dialectological scenario in which a variable form is centered on one specific city, as in, say, the association of the word \ex{hella} with the San Francisco metropolitan area. We average across twenty five possible centers: the capitals of each of the twelve provinces of the Netherlands; the national capital of the Netherlands (Amsterdam); the \emph{two} most populous cities in each of the twelve provinces. For each setting, we randomly generate synthetic data four times, resulting in a total of 100 synthetic datasets for this condition.
\end{description}

\begin{figure}
\centering
\begin{subfigure}{0.4\textwidth}
  \centering
  \includegraphics[width=0.74\linewidth]{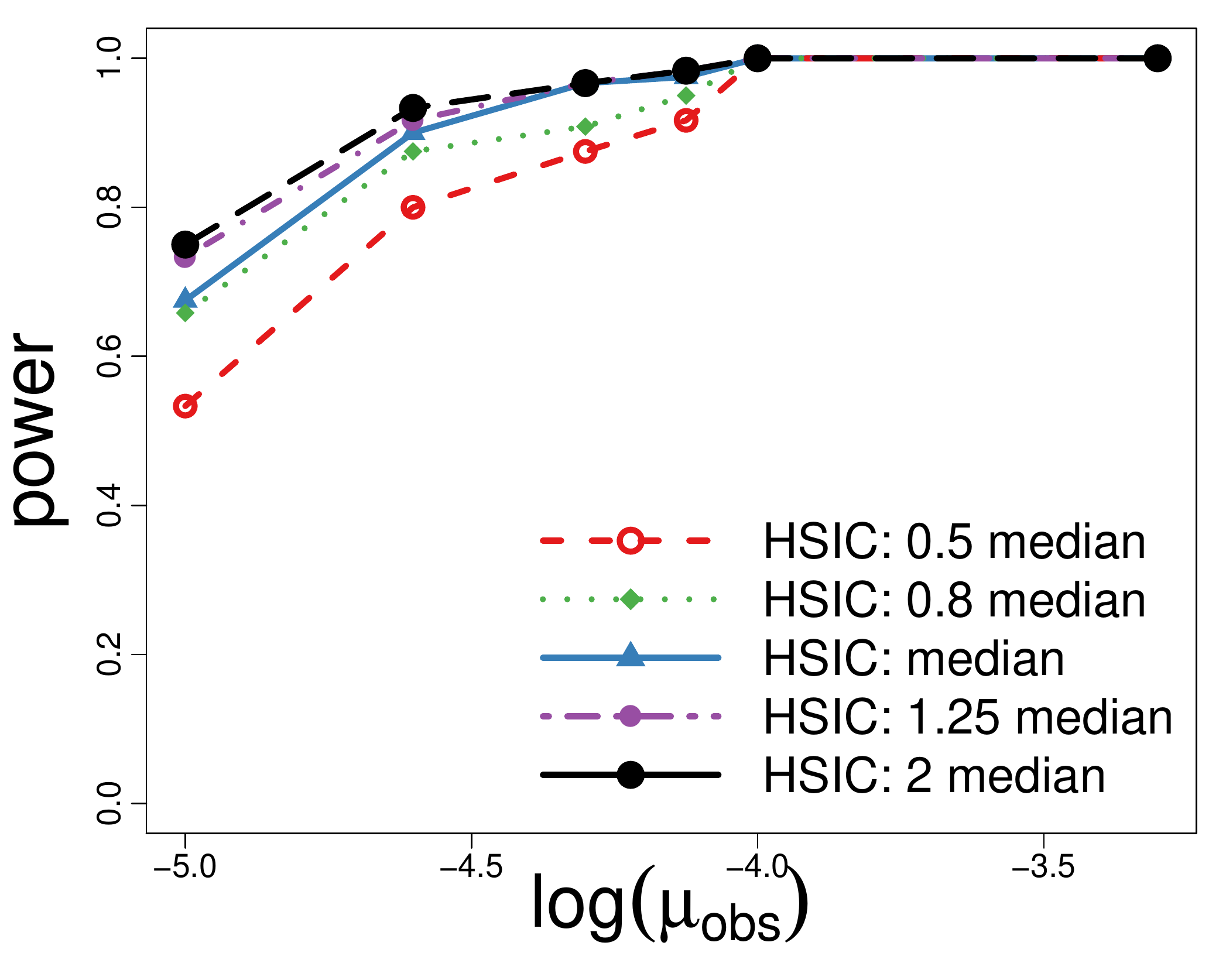}
  \caption{HSIC: Continua}
  \label{fig:hsic-power-param-linear}
\end{subfigure}%
\begin{subfigure}{.4\textwidth}
  \centering
  \includegraphics[width=0.74\linewidth]{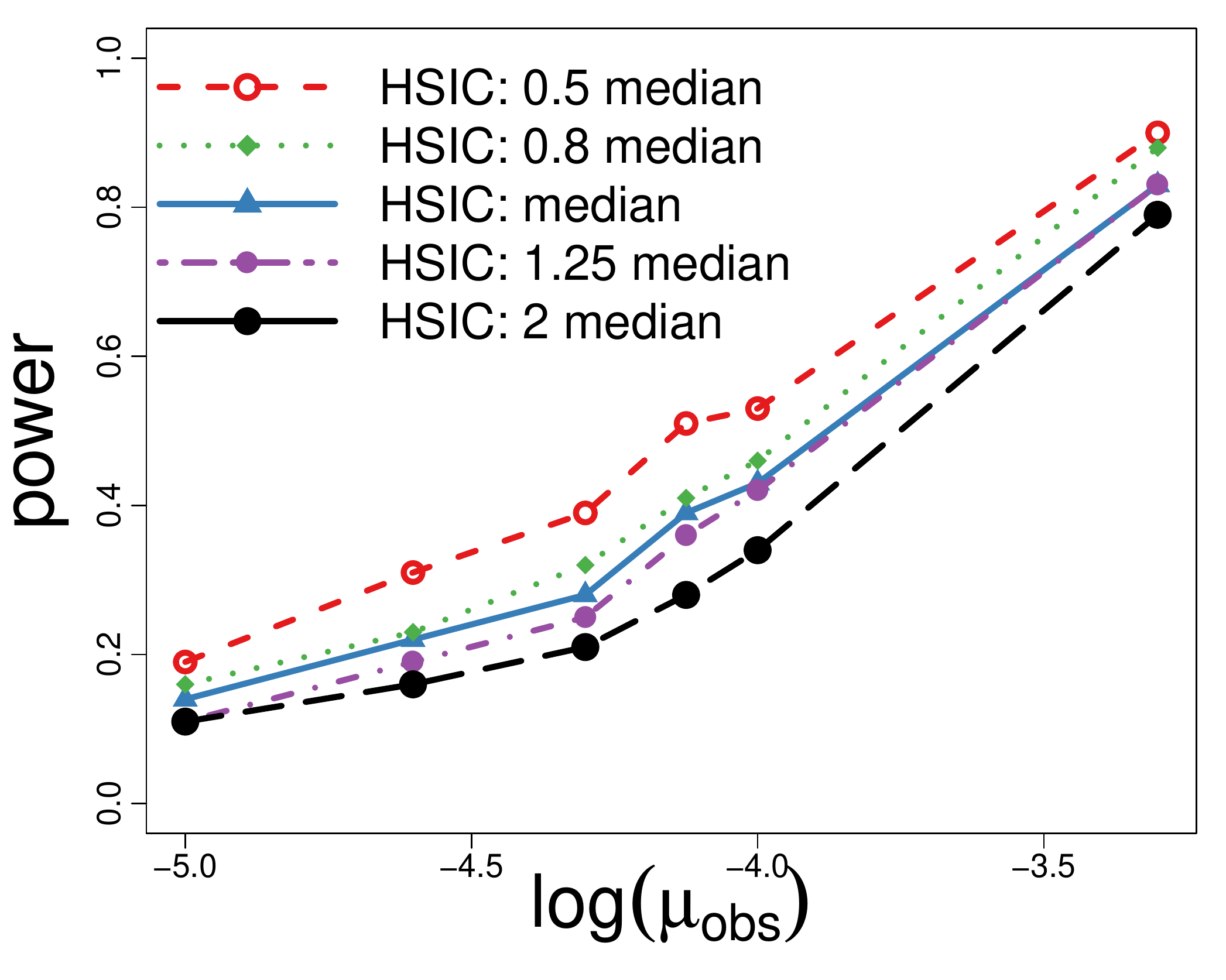}
  \caption{HSIC: Centers}
  \label{fig:hsic-power-param-center}
\end{subfigure}
\begin{subfigure}{0.4\textwidth}
  \centering
  \includegraphics[width=0.74\linewidth]{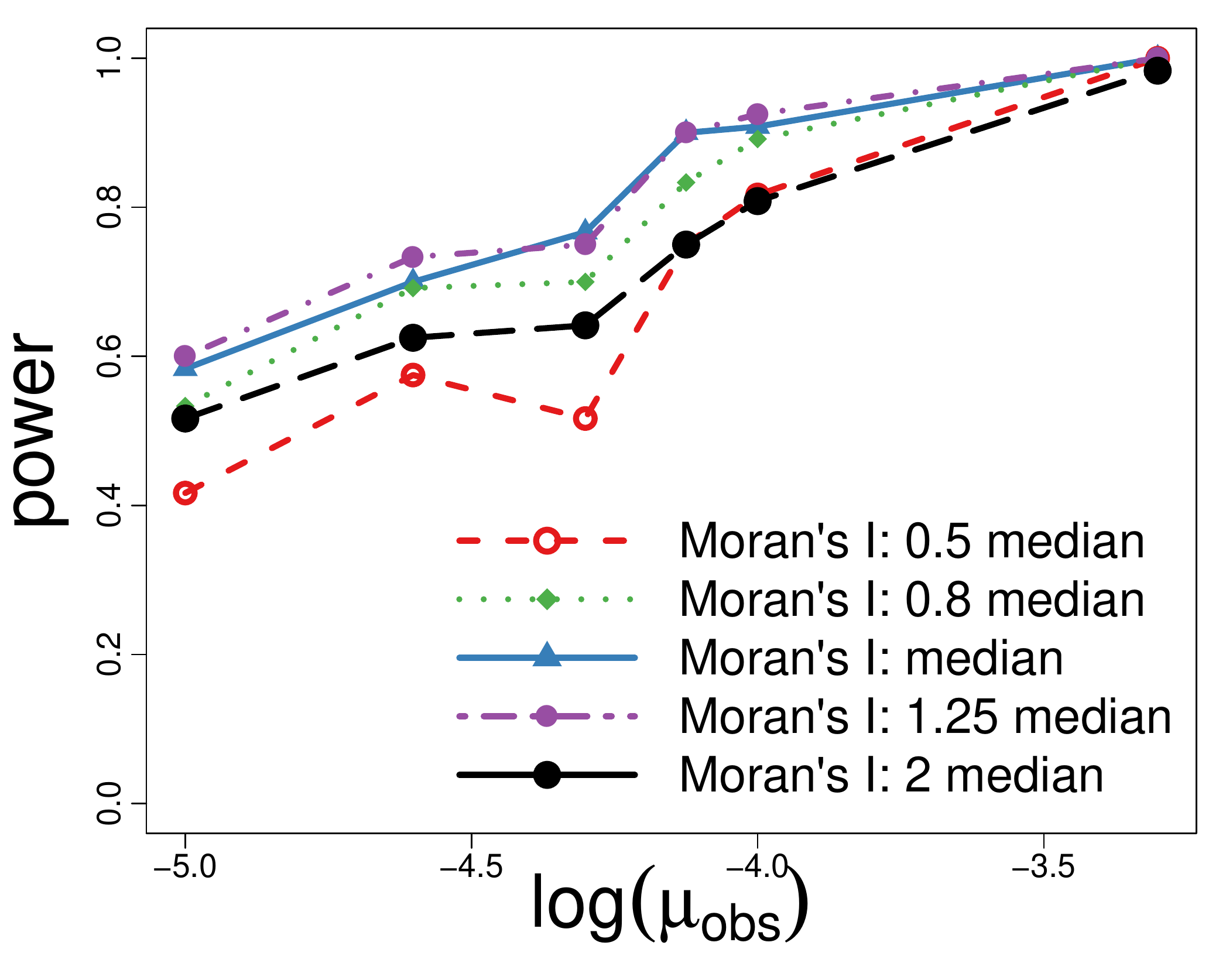}
  \caption{\mi: Continua}
  \label{fig:mi-power-param-linear}
\end{subfigure}%
\begin{subfigure}{.4\textwidth}
  \centering
  \includegraphics[width=0.74\linewidth]{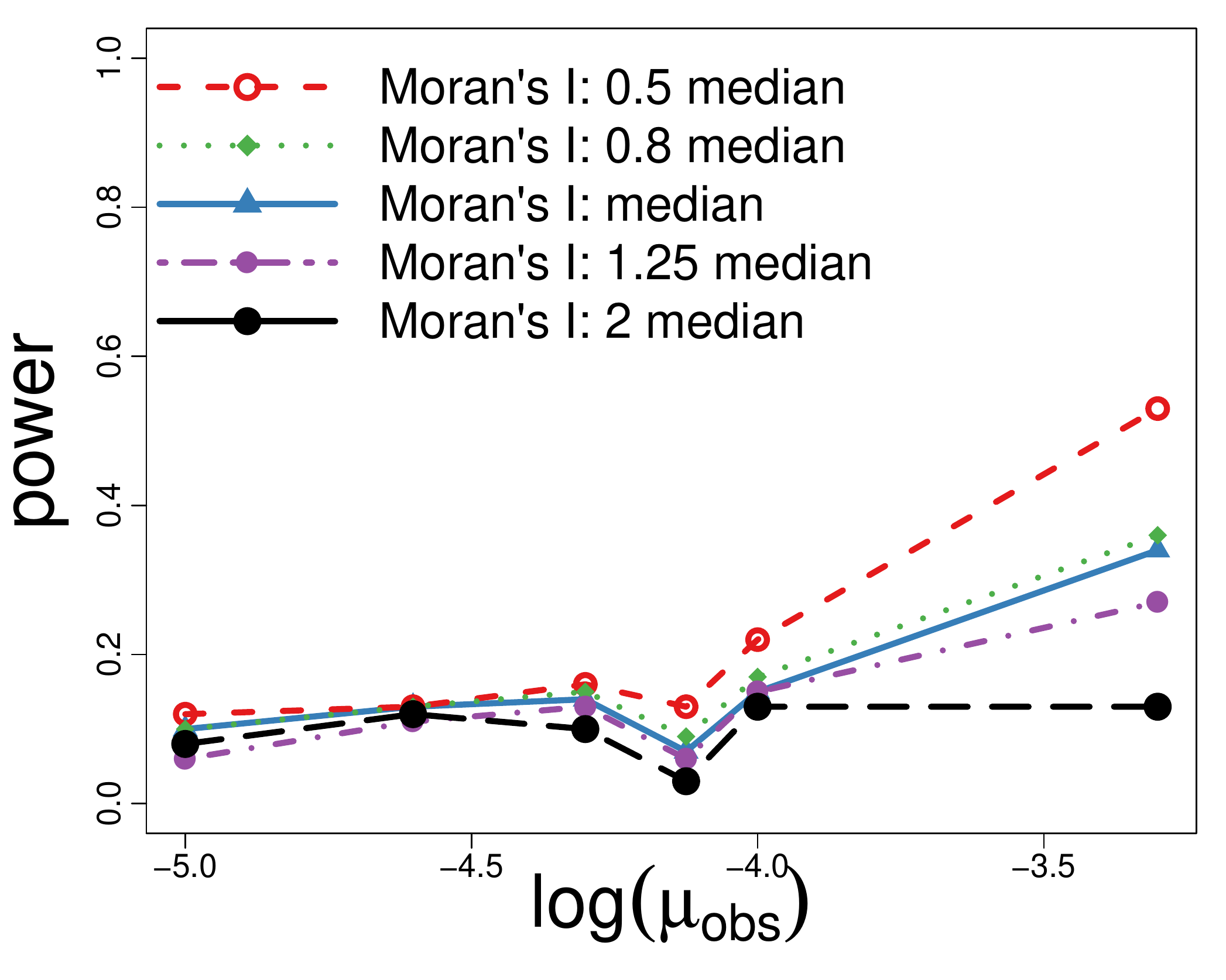}
  \caption{\mi: Centers}
  \label{fig:mi-power-param-center}
\end{subfigure}
\caption{Power across different parameter settings. Higher values indicate a greater likelihood of correctly rejecting the null hypothesis.}
\label{fig:power-params}
\end{figure}

\paragraph{Parameter settings}  
We use these data generation scenarios to test the sensitivity of \hsic\ and \mi\ to their hyperparameters, by varying the kernel bandwidths in \hsic\  (Figures~\ref{fig:hsic-power-param-linear} and \ref{fig:hsic-power-param-center}) and the distance threshold in \mi\ (Figures~\ref{fig:mi-power-param-linear} and \ref{fig:mi-power-param-center}). The sensitivity of \hsic\ to the bandwidth value decreases as the number of data points increases  (as governed by $\mu_{obs}$), especially in the case of \new{dialect continua}. The sensitivity of \mi\ to the distance cutoff value decreases with the amount of data in the case of dialect continua, but in the case of center-based variation, \mi\ becomes \emph{more} sensitive to this parameter as there is more data. For both methods, the same trends regarding the best performing parameters can be observed. In the case of dialect continua, larger cutoffs and bandwidths perform best, but in the case of variation based on centers, smaller cutoffs and bandwidths lead to higher power. Overall, there is no single best parameter setting, but the median heuristics perform reasonably well for both types of variation.

\begin{figure}
\centering
  \includegraphics[width=.32\linewidth]{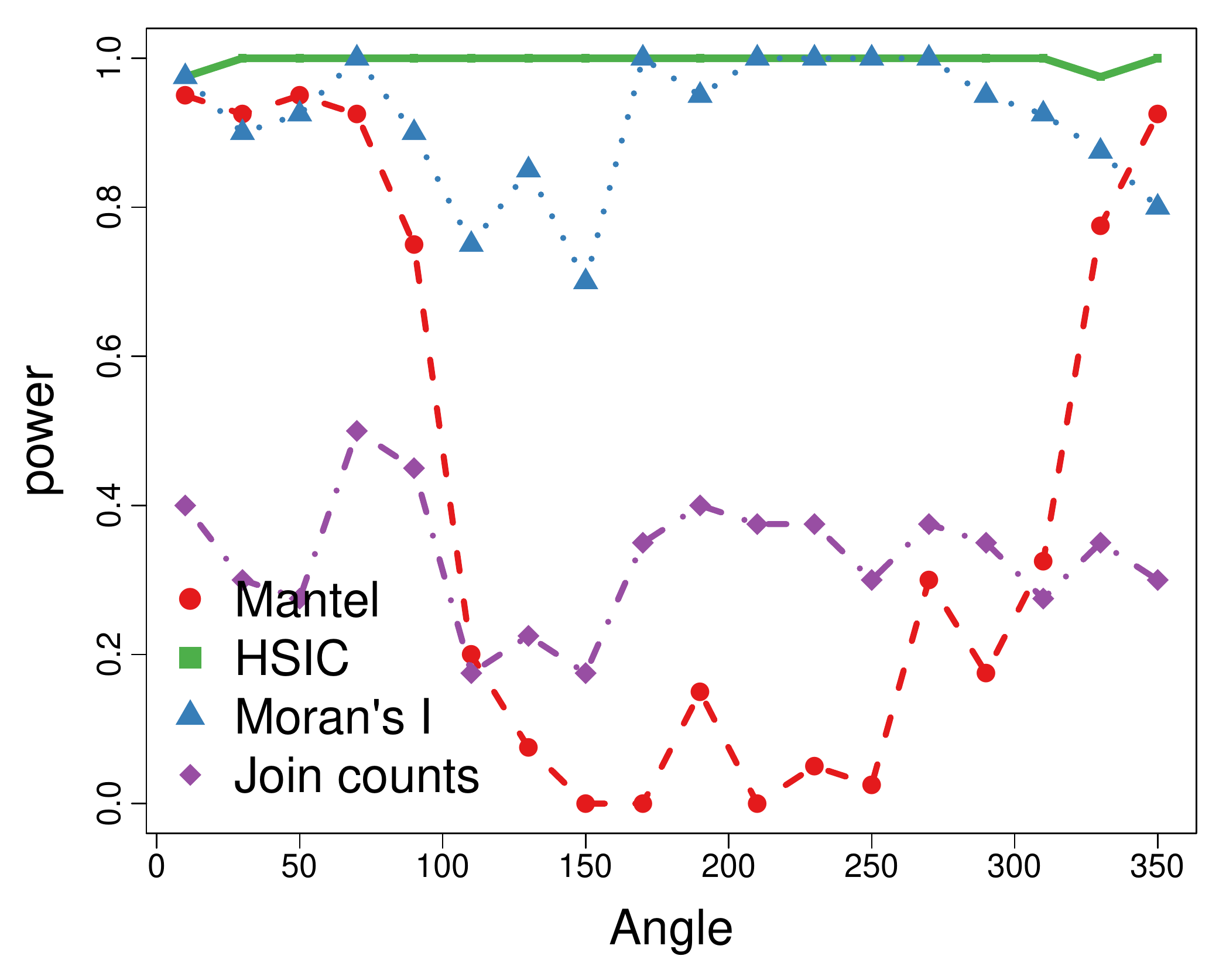}
  \caption{Relationship between the statistical power of each test and the angle of the dialect continuum across the Netherlands.}
\label{fig:2cat_angle_linear}
\end{figure}

\paragraph{Direction of dialect continua} 
We simulate dialect continua by varying the frequency of linguistic variables linearly through space. Due to the heterogeneity of population density, different spatial angles will have very different properties: for example, one choice of angle would imply a continuum cutting through several major cities, while another choice might imply a rural-urban distinction. \autoref{fig:2cat_angle_linear} shows the power of the methods on binary data (there are two variant forms, and each instance contains exactly one of them), in which we vary the angle of the continuum. \hsic\ is insensitive to the angle of variation, demonstrating the advantage of this kernel nonparametric method. \mi\ is relatively robust, while join count analysis performs poorly across the entire range of settings. The \mt\ is remarkably sensitive to the angle of variation, attaining nearly zero power for some scenarios of dialect continua. 
This is caused by the complex interaction between the underlying linguistic phenomenon and the east-west variation of the population density of the Netherlands. For example, when the dialect continuum is simulated at an angle of 105 degrees, the south east of the Netherlands has a higher usage of the variable, but this is only a very small region due to the shape of the country. The \mt\ apparently has great difficulty in detecting geographical variation in such cases.



\begin{figure}
\centering
\begin{subfigure}{0.35\textwidth}
  \centering
  \includegraphics[width=0.9\linewidth]{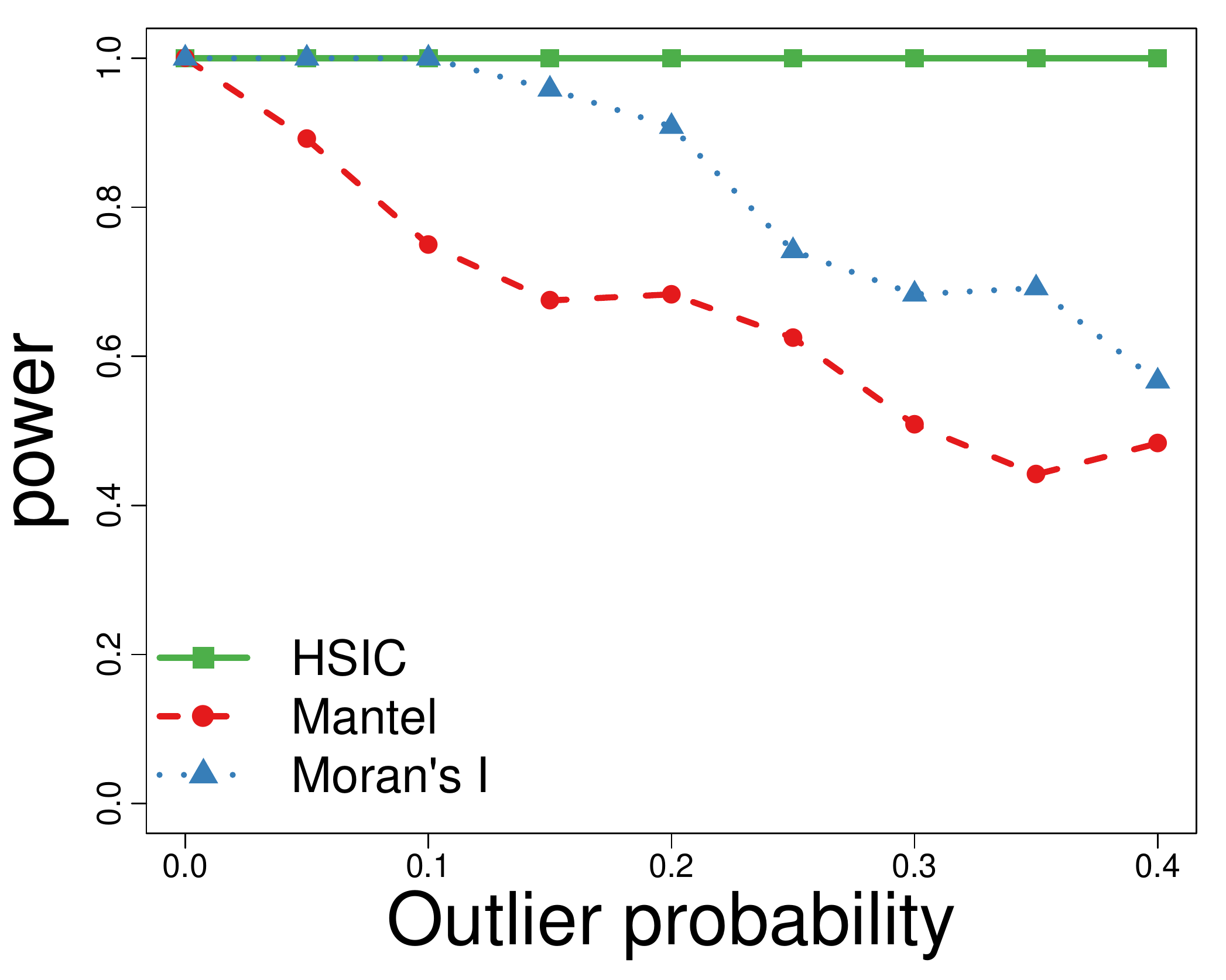}
  \caption{Dialect continua}
\end{subfigure}%
\begin{subfigure}{0.35\textwidth}
  \centering
  \includegraphics[width=0.9\linewidth]{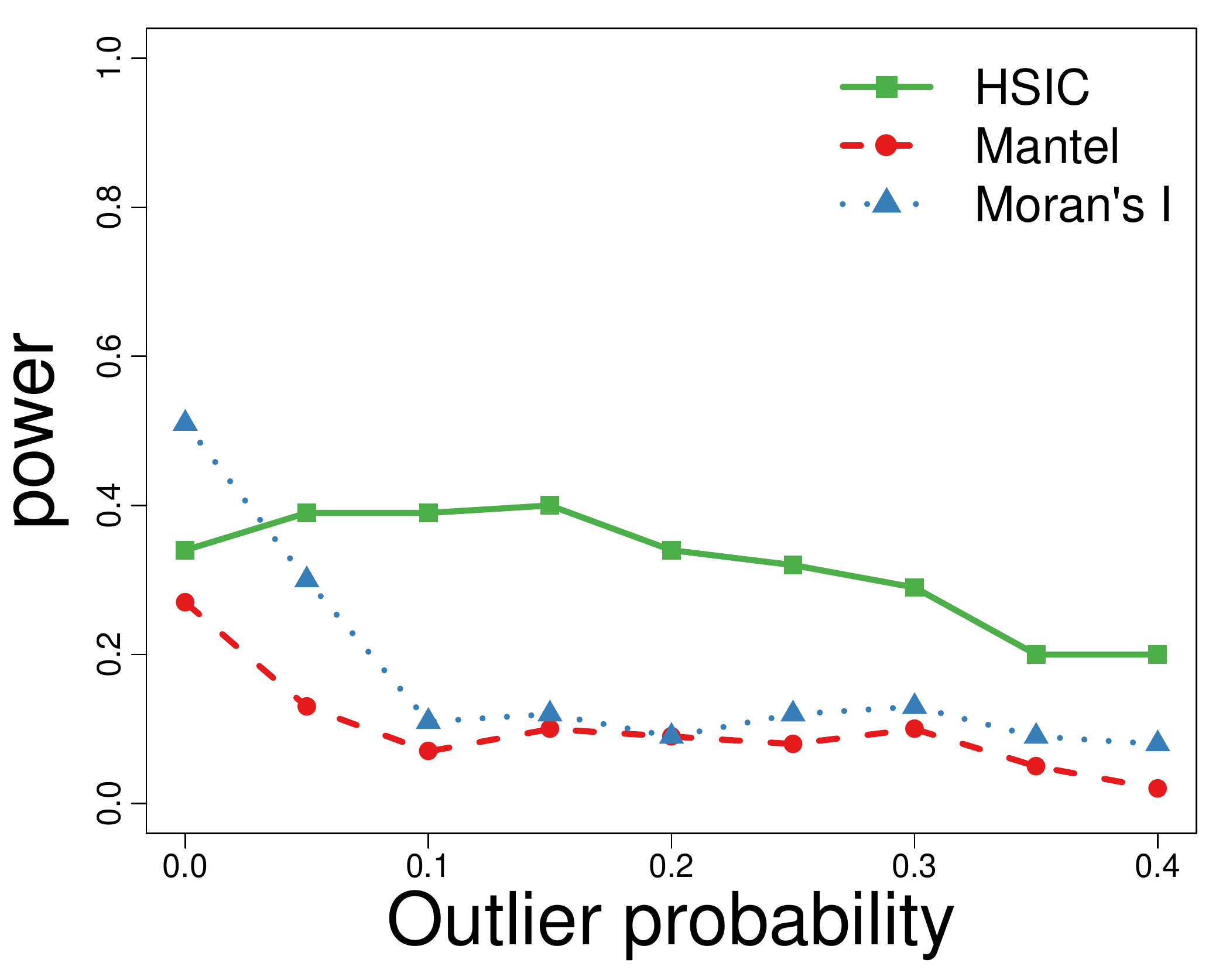}
  \caption{Centers}
\end{subfigure}

\caption{Results on synthetic frequency data ($\sigma=0.1$) with outliers}
\label{fig:synthetic_freq_noise}
\end{figure}

\paragraph{Outliers} 
In the frequency-based synthetic data, each instance uses each variable form with some continuous frequency --- this is based on the scenario of letters-to-the-editors of regional newspapers, as explored in prior work~\cite{grieve2011statistical}. We test the robustness of each approach by introducing \emph{outliers}: randomly selected data points whose variable frequencies are replaced at random with extreme values of either $0$ or $1$. As shown in \autoref{fig:synthetic_freq_noise}, \hsic\ is the most robust against outliers, while the performance of \new{the Mantel test} is the most affected by outliers (recall that join count analysis applies only to discrete observations, so it cannot be compared on this measure).
 
\begin{figure}
\centering
\begin{subfigure}{0.3\textwidth}
  \centering
  \includegraphics[width=0.9\linewidth]{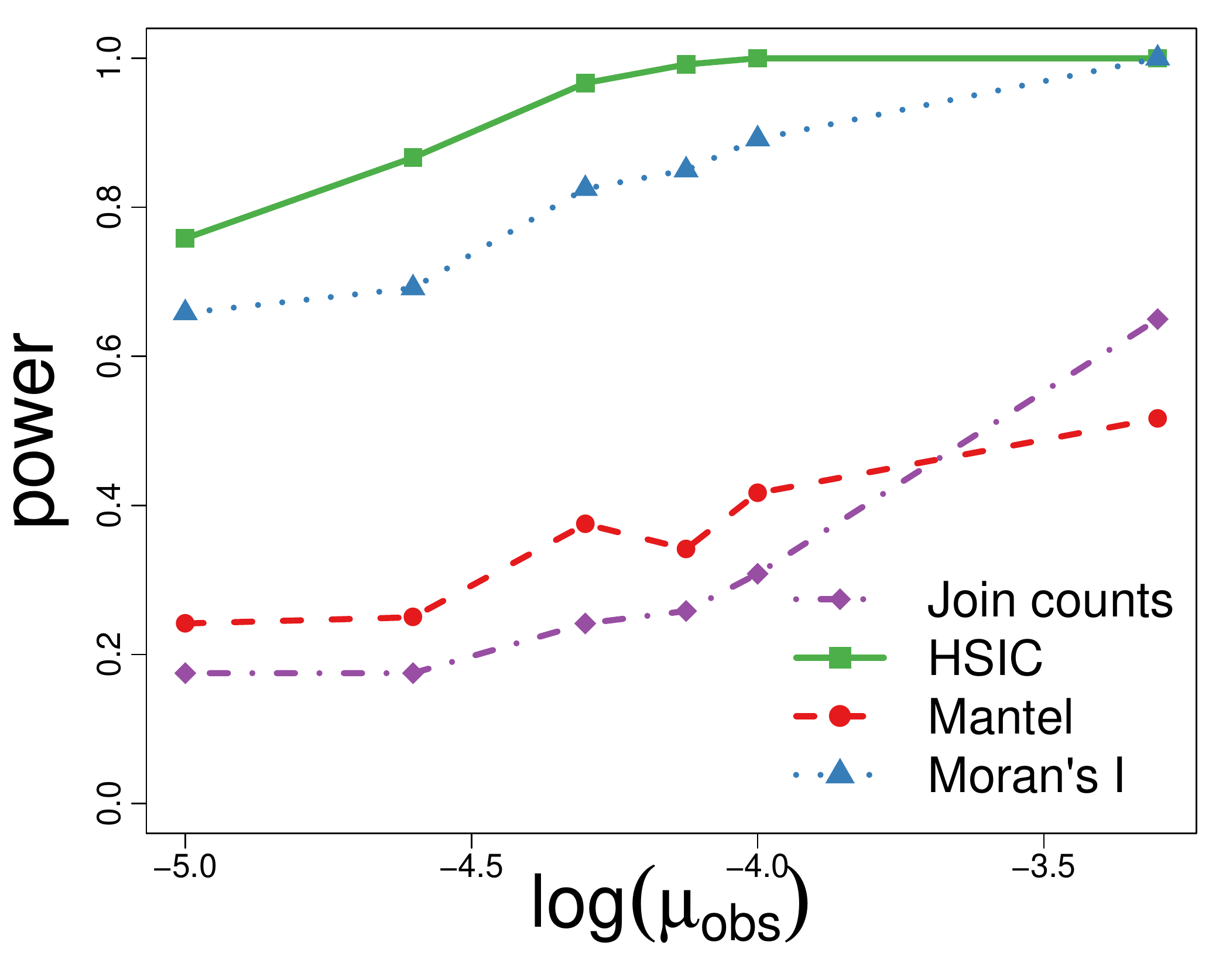}
  \caption{Binary data}
\end{subfigure}%
\begin{subfigure}{.3\textwidth}
  \centering
  \includegraphics[width=0.9\linewidth]{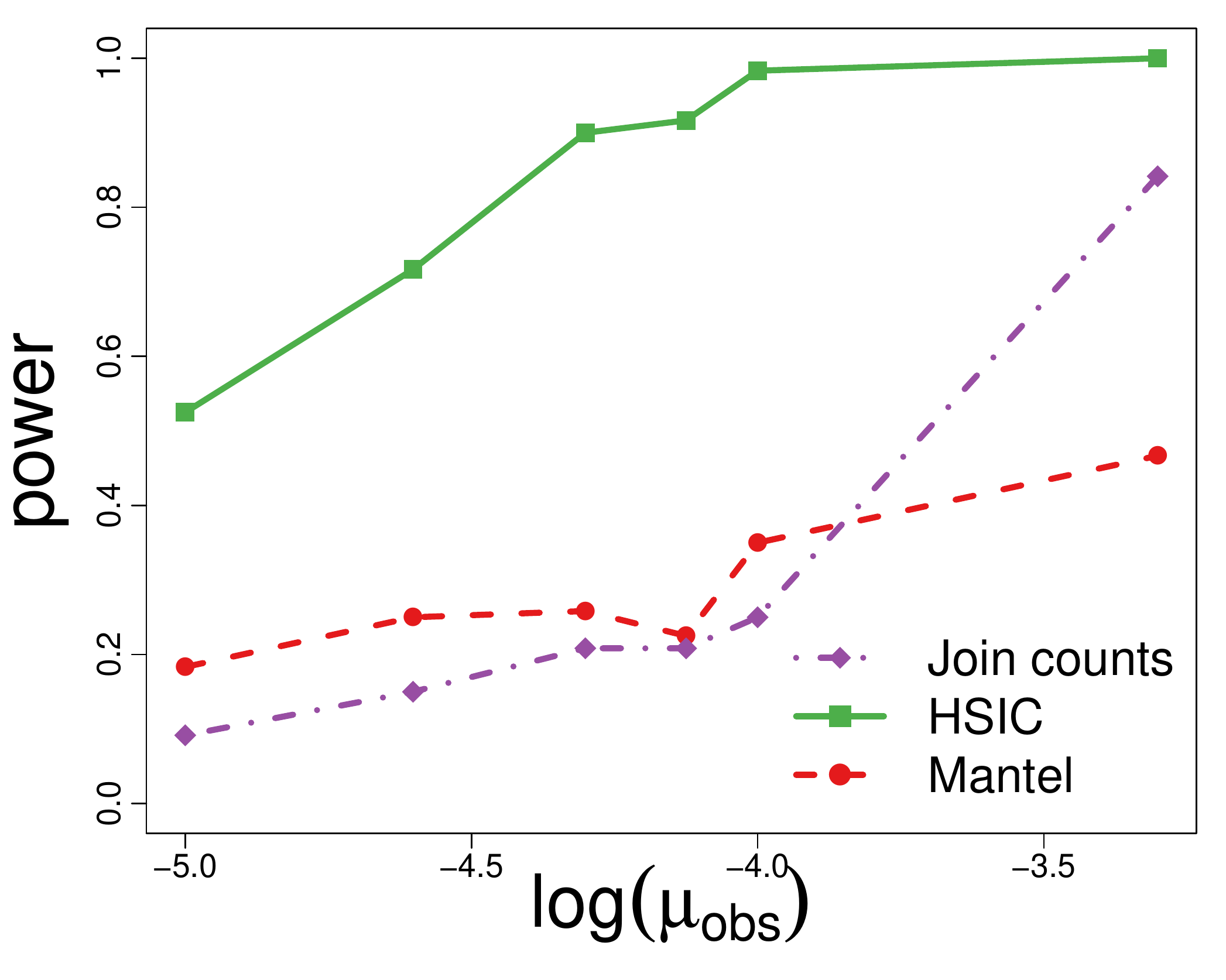}
  \caption{Categorical data (3)}
\end{subfigure}
\begin{subfigure}{.3\textwidth}
  \centering
  \includegraphics[width=0.9\linewidth]{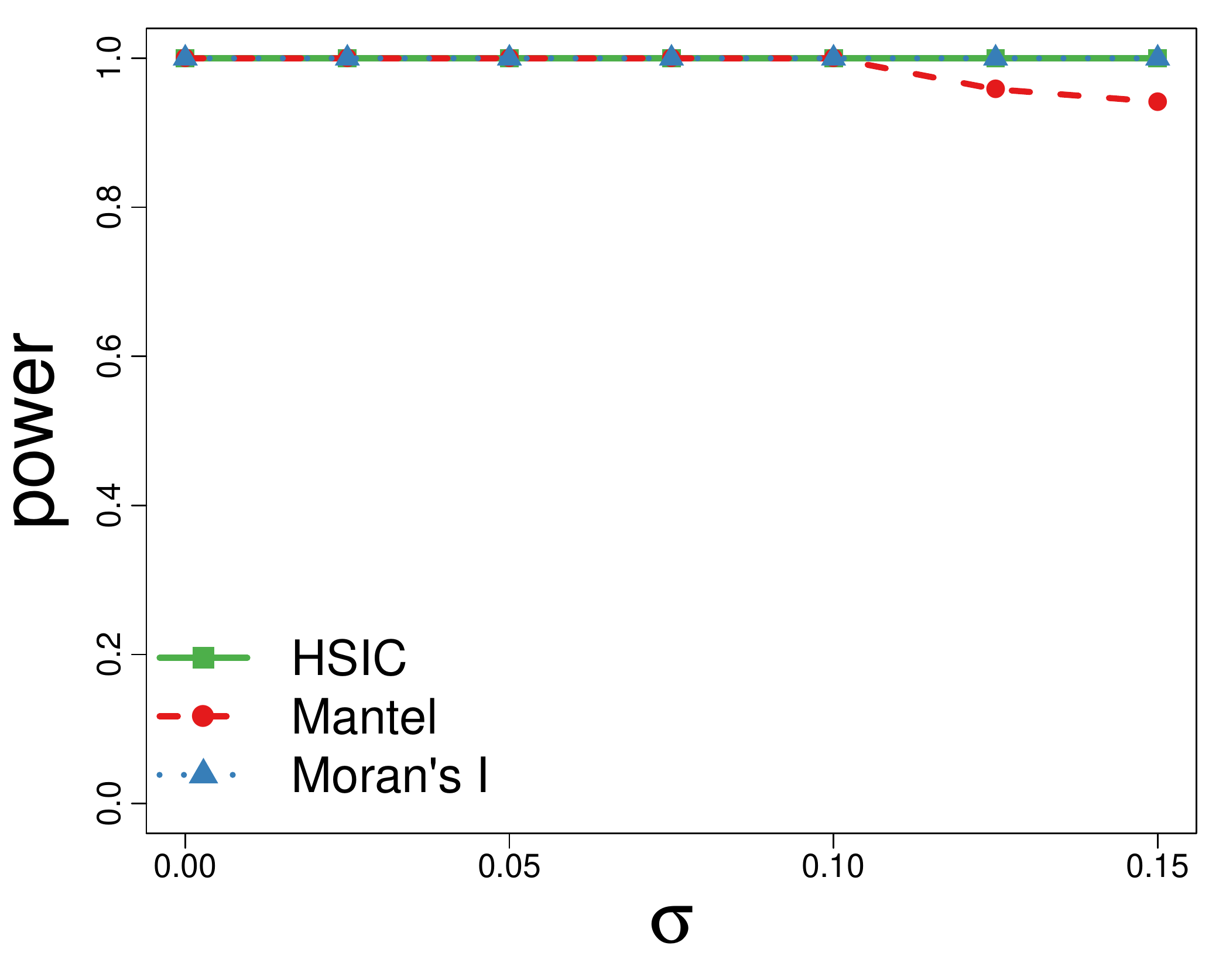}
  \caption{Frequency data}
\end{subfigure}
\caption{Dialect continua}
\label{fig:linear_average}
\end{figure}

\begin{figure}
\centering
\begin{subfigure}{0.3\textwidth}
  \centering
  \includegraphics[width=0.9\linewidth]{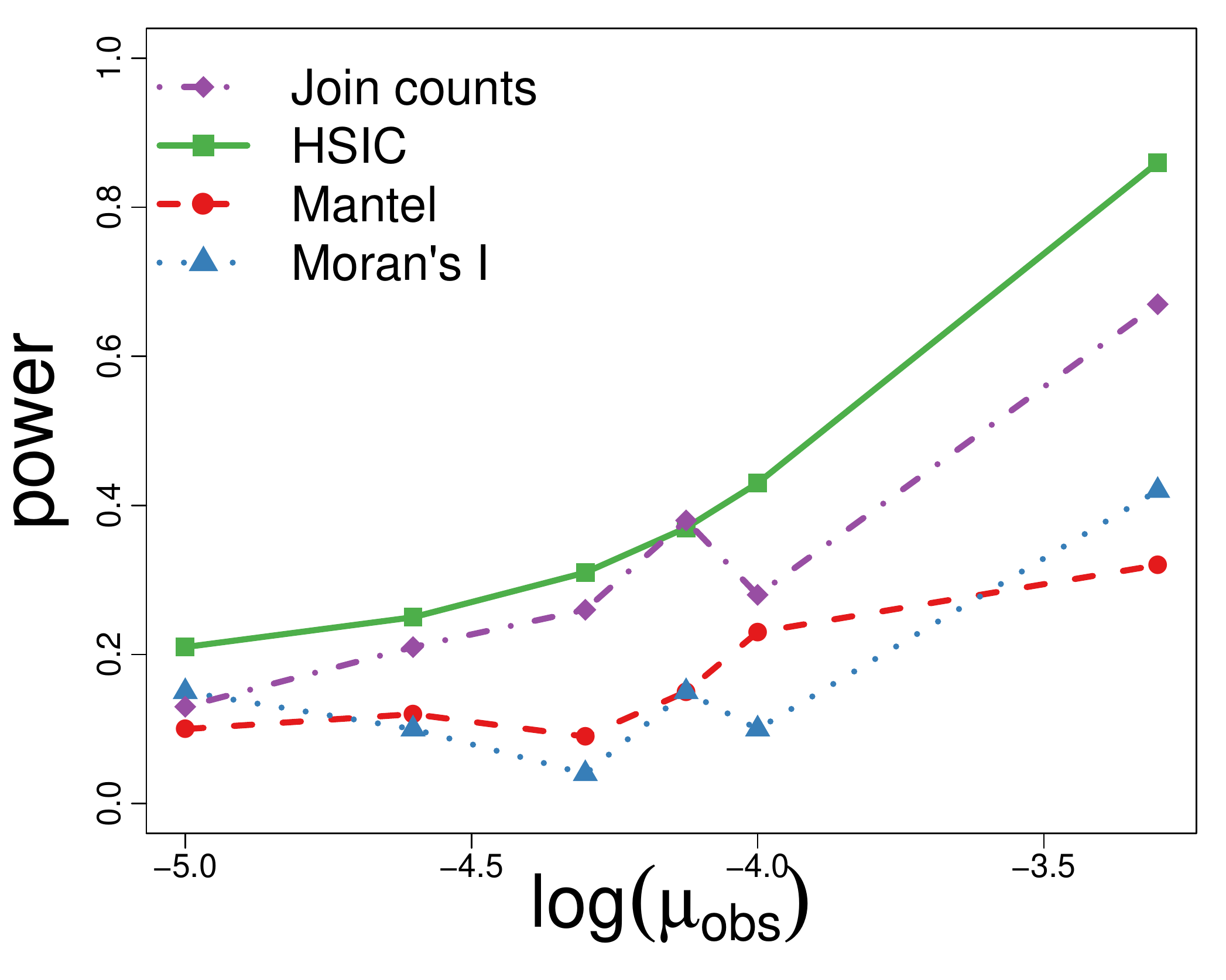}
  \caption{Binary data}
\end{subfigure}%
\begin{subfigure}{.3\textwidth}
  \centering
  \includegraphics[width=0.9\linewidth]{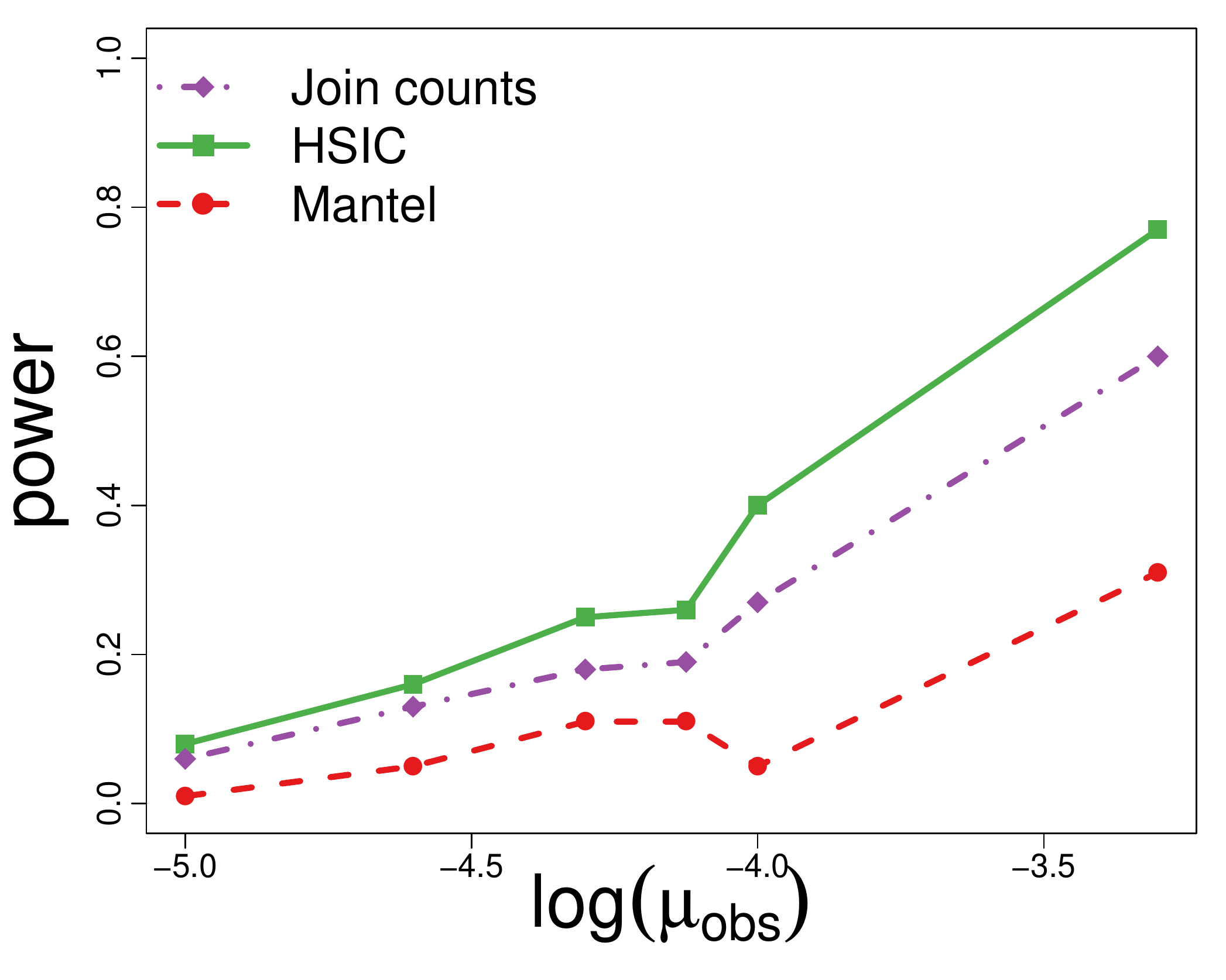}
  \caption{Categorical data (3)}
\end{subfigure}
\begin{subfigure}{.3\textwidth}
  \centering
  \includegraphics[width=0.9\linewidth]{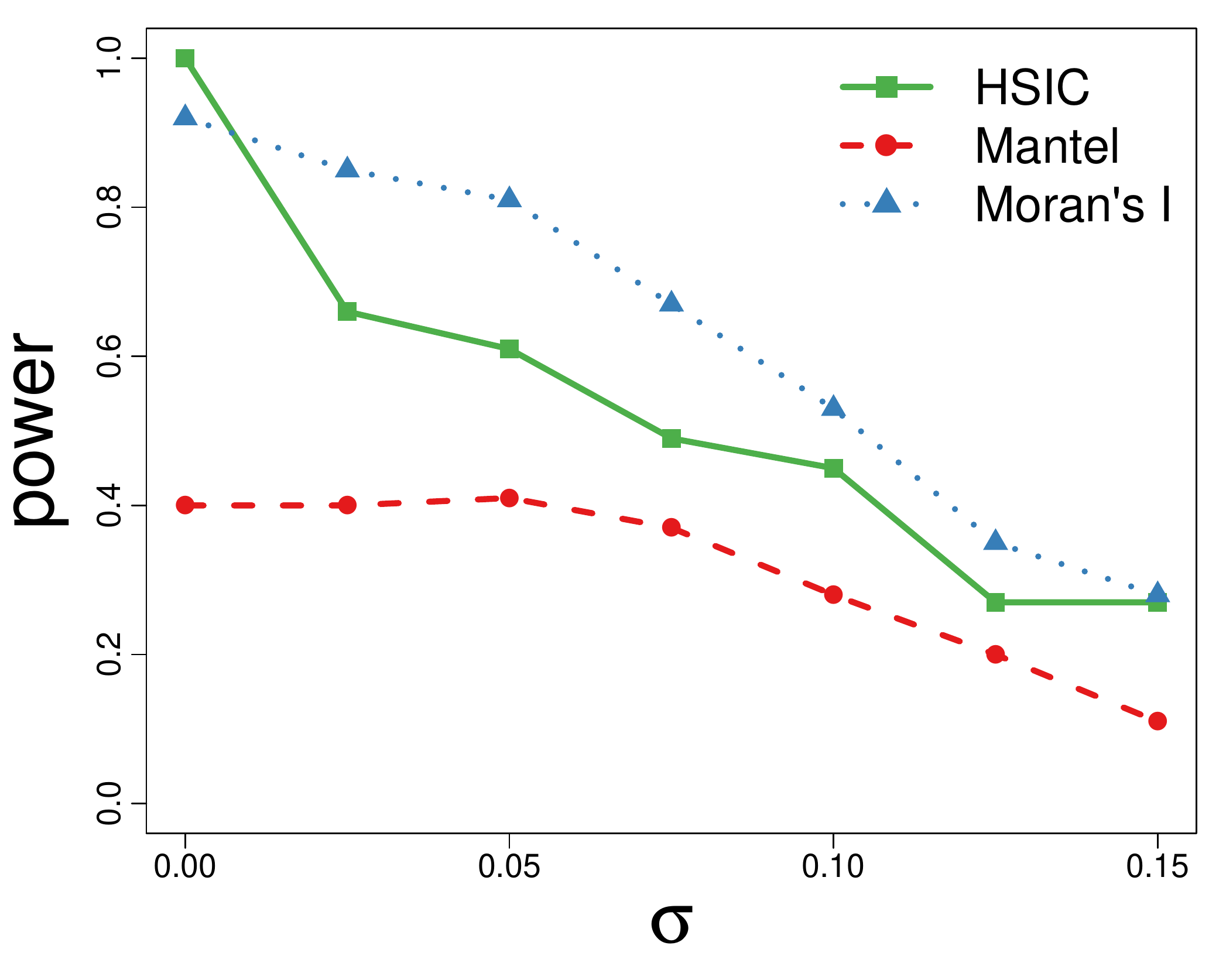}
  \caption{Frequency data}
\end{subfigure}
\caption{Centers}
\label{fig:centers_average}
\end{figure}

\paragraph{Overall}
We now compare the methods by averaging across various settings simulating dialect continua (Figure \ref{fig:linear_average})
and variation based on centers (Figure \ref{fig:centers_average}).
To generate the categorical data, we vary $\mu_{obs}$ in our experiments, with a higher $\mu_{obs}$ resulting in more tweets and consequently less variation on the municipality level. 
As expected, the power of the approaches increases as $\mu_{obs}$ increases in the experiments on the categorical data, and the power of the approaches decreases as $\sigma$ increases in the experiments on  the frequency data. The experiments on the binary and categorical data show the same trends: \hsic\ performs the best across all settings. Join count analysis does well when the variation is based on centers, and \mi\ does best for dialect continua. \mi\ performs best on the frequency data, especially in the case of variation based on centers. 

\subsection{Summary}
In this section, we evaluated each statistical test for geographical language variation on a battery of synthetic data. 
\hsic\ and the Mantel test are the only approaches applicable to all data types (binary, categorical and frequency data). Overall, \hsic\ is more effective than the Mantel test, which is much more sensitive to the specifics of the synthetic data scenario, such as the angle of the dialect continuum. \hsic\ is robust against outliers, and performs particularly well when the number of data points increases. \new{Join count analysis} is suitable for capturing non-linear variation, but its power is low compared to other approaches in the analysis of dialect continua. Conversely, when the linguistic data is binary, \mi\ performs well on dialect continua, but its power is low in situations of variation based on centers. In our experiments on frequency data, \mi\ performs well in both scenarios.

%% file: empirical.tex
\mynewpage
\section{Empirical Data}
\label{sec:real}
We now assess the spatial dependence of linguistic variables on three real linguistic datasets: letters to the editor (English), syntactic atlas of the Dutch dialects, and Dutch geotagged tweets. In each dataset, we compute statistical significance for the geo-linguistic dependence of multiple linguistic variables. \new{To adjust the significance thresholds for multiple comparisons, we use the Benjamini-Hochberg procedure~\cite{Benjamini1995} to bound the overall false discovery rate (FDR).}

\subsection{Letters to the Editor}
In their application of \mi\ to English dialects in the United
States, \namecite{grieve2011statistical} compile a corpus of letters-to-the-editors of
newspapers to measure the presence of dialect variables  in text.  To
compute the frequency of the lexical variables, letters are
aggregated to core-based statistical areas (CBSA), which are
defined by the United States to capture the geographical region around
an urban core. The frequency of 40 manually selected lexical variables is computed for each of 206 cities. 

We use the \mt, \hsic, and \mi\ to assess the spatial dependence of variables in this dataset. \new{Join count analysis} was excluded, because it is not suitable for frequency data. We verified our implementation of \mi\ by following the approach taken by Grieve \etal: we computed \mi\ for cutoffs in the range of 200 to 1000 miles and selected the cutoff that yielded the lowest $p$-value. The obtained cutoffs and test statistics closely followed the values reported in the analysis by Grieve \etal, with slight deviations possibly due to our use of a permutation test rather than a closed-form approximation to compute the $p$-values.

\begin{figure}
 \includegraphics[scale=0.23]{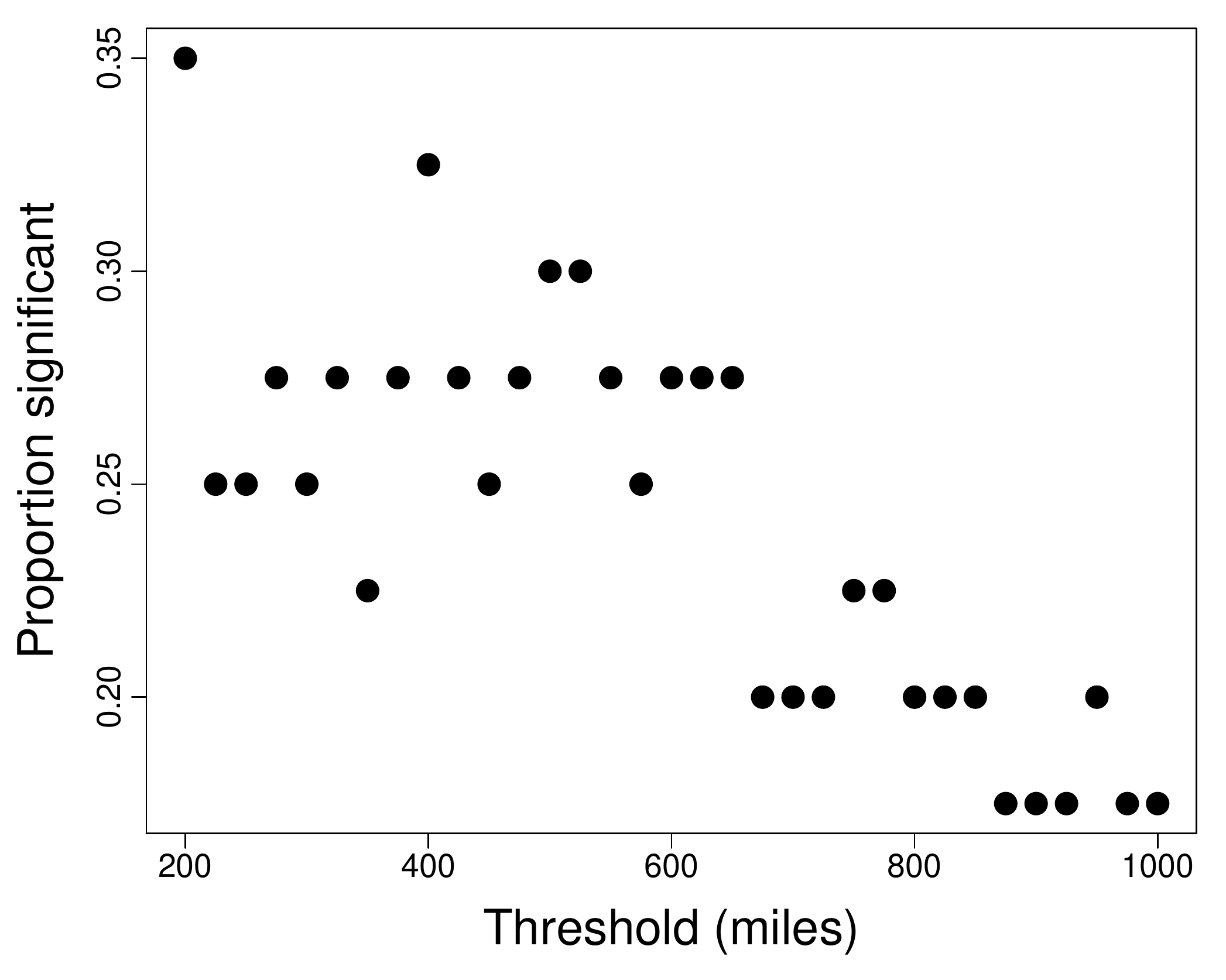}
 \caption{The proportion of variables detected to be significant ($p$<.05) by Moran's I
 by varying the distance cutoff (without  adjusting for multiple comparisons).}
 \label{lvc_thresholds}
\end{figure}

After adjusting the $p$-values using the false discovery rate (FDR) procedure, a 500-mile cutoff results in three significant linguistic variables.\footnote{Grieve \etal\ report five significant variables. In our analysis, there are two variables with FDR-adjusted $p$-values of 0.0559} However, recall that the approach of selecting parameters by maximizing the number of positive test results tends to produce a large number of Type I errors. When setting the distance cutoff to the median distance between data points, none of the linguistic variables were found to have a significant geographical association. Similarly, \hsic\ and the Mantel test also found no significant associations after adjusting for multiple comparisons. Figure \ref{lvc_thresholds} shows the proportion of significant variables according to \mi\ based on different thresholds. The numbers vary considerably depending on the threshold. The figure also suggests that the median distance (921 miles) may not be a suitable threshold for this dataset.

\subsection{Syntactic Atlas of the Dutch Dialects (SAND)}
SAND \cite{barbiers2005syntactic,barbiers2008} is an online electronic atlas that maps syntactic variation of Dutch varieties in the Netherlands, Belgium, and France.\footnote{http://www.meertens.knaw.nl/sand/} The data was collected between the years of 2000 and 2005. SAND has been used to measure the distances between dialects, and to discover dialect regions \cite{Spruit01112006,tks2015edisyn}.

In our experiments, we consider only locations within the Netherlands (157 locations). The number of variants per linguistic variable ranges from one (due to our restriction to the Netherlands) to eleven.
We do not include \mi\ in our experiments, since it is not applicable to linguistic variables with more than two variants. We apply the remaining methods to all linguistic variables with twenty or more data points and at least two variants, resulting in a total of 143 variables.

\begin{figure}
   \CenterFloatBoxes
\begin{floatrow}

\capbtabbox{%
    \centering
    \small
\begin{tabular}{llrrr}
\toprule
\textbf{Map id} & \textbf{Description}\\
\midrule
1:84b & \specialcell[t]{Free relative, complementizer \\following relative pronoun} \\
1:84a &  \specialcell[t]{Short subject and object relative, \\complementizer following relative\\ pronoun}\\ 
1:80b &ONE pronominalisation\\
1:33a & Complementizer agreement 3 plural \\
1:76a & Reflexive pronouns; synthesis \\
2:36b & \specialcell[t]{ Form of the participle of \\the modal verb willen `want'}\\
1:29a & Complementizer agreement 1 plural\\
1:69a & Correlation weak reflexive pronouns\\
2:30b & \specialcell[t]{Interruption of the verbal cluster;\\ synthesis I}\\
2:61a & Forms for iemand `somebody'\\
\bottomrule
\end{tabular}
}{%
\caption{Highest ranked variables by HSIC. All methods had an adjusted $p$-value $<.05$ for all variables,
except variable 2:30b (Mantel: $p=.118$.}
\label{tab:sand_hsic_top}
}
\ffigbox{%
 \includegraphics[scale=0.3]{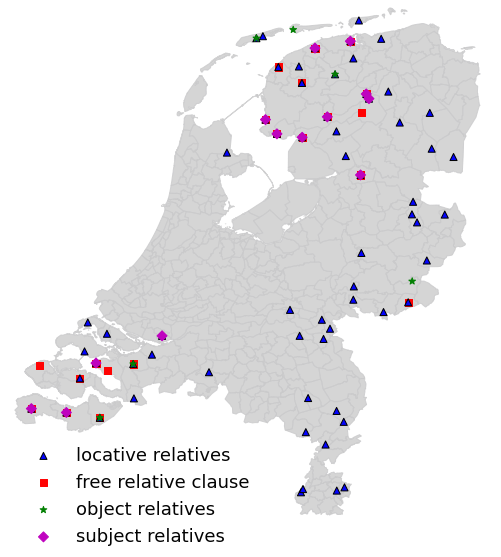}
}{%
 \caption{SAND map 16b, book 1. \\Finite complementizer(s) following \\relative pronoun (N=112).\\
  HSIC: $p$=.024; Mantel: $p$=.506;\\ Join counts: $p$=.002.
 }.

 \label{fig:sand_mantel_example}
}
\end{floatrow}
\end{figure}

Table \ref{tab:sand_hsic_top} lists the 10 variables with the highest \hsic\ values. Statistical significance at a level of $\alpha=0.05$ is detected for 65.0\% of the linguistic variables using \hsic, 78.3\% when using \new{join count analysis}, and 52.4\% when using the Mantel test. The three methods agree on 99 out of the 143 variables, and \hsic\ and join count analysis agree on 118 variables.  
From manual inspection, it seems that the non-linearity of the geographical patterns may have caused difficulties for the \mt. For example, \autoref{fig:sand_mantel_example} is an example of a variable where \hsic\ and \new{join count analysis} both had an adjusted $p$-value $<$ .05, but the Mantel test did not detect a significant pattern. 

\mynewpage
\subsection{Twitter}
\label{sec:twitter}
Our Twitter dataset consists of 4,039,786 geotagged tweets from the Netherlands, written between January 1, 2015 and October 31, 2015. 
We manually selected a set of linguistic variables (\autoref{tab:results_twitter}), covering examples of lexical variation (e.g., two different words for referring to french fries),
phonological variation (e.g., t-deletion), and syntactic variation (e.g., \ex{heb gedaan} (`have done') vs. \ex{gedaan heb} (`done have')). We are not aware of any previous work on dialectal variation in the Netherlands that uses spatial dependency testing on Twitter data. The number of tweets per municipality varies dramatically, and for the less frequent linguistic variables there are no tweets at all in some municipalities. In our computation of \mi, we only include municipalities with at least one tweet.

\begin{figure}
\centering
\begin{subfigure}{0.4\textwidth}
  \centering
  \includegraphics[width=1\linewidth]{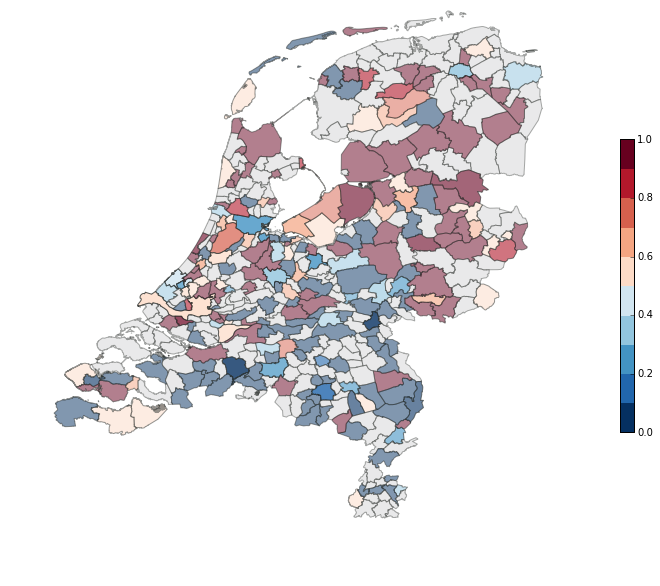}
  \caption{French fries (\emph{friet} versus \emph{patat}), N=844}
  \label{fig:friet}
\end{subfigure}%
\hspace{0.2in}
\begin{subfigure}{.4\textwidth}
  \centering
  \includegraphics[width=1\linewidth]{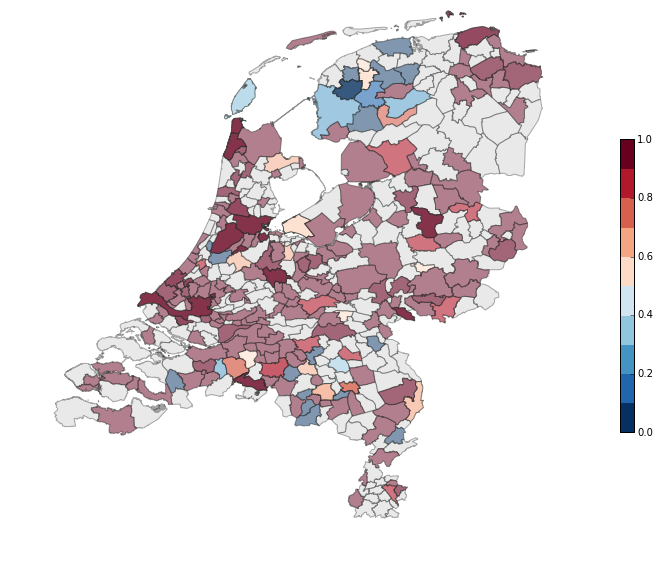}
  \caption{For a little while (\emph{efkes} versus \emph{eventjes}), N=970}
  \label{fig:efkes}
\end{subfigure}

\caption{Highly significant linguistic variables on Twitter. Grey indicates areas with no data points.  The intensity indicates the number of data points.}
\label{fig:twitter_sign}
\end{figure}

\autoref{tab:results_twitter} shows the output of each statistical test for this data. Some of these linguistic variables
exhibit strong spatial variation, and are identified as statistically significant by all approaches. An example is the different ways of referring to french fries (\ex{friet} versus \ex{patat}, \autoref{fig:friet}), where the figure shows a striking difference between the south and the north of the Netherlands. 
Another example is \autoref{fig:efkes}, which shows two different ways of saying `for a little while' (\ex{efkes} versus \ex{eventjes}). The less common form, \ex{efkes} is mostly used in Friesland, a province in the north of the Netherlands. 

\begin{figure}
\centering
\begin{subfigure}{0.4\textwidth}
  \centering
  \includegraphics[width=1\linewidth]{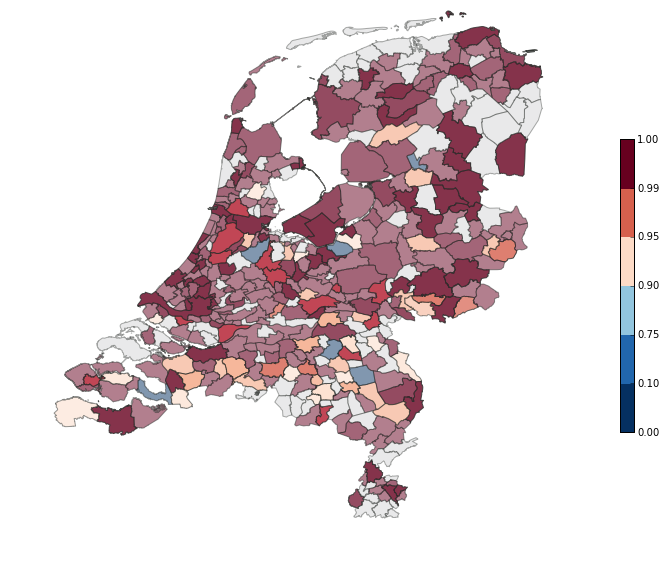}
  \caption{Bye (\ex{doei} versus \ex{houdoe}), N=4,491}
  \label{fig:doei_houdoe}
\end{subfigure}%
\hspace{0.2in}
\begin{subfigure}{.4\textwidth}
  \centering
  \includegraphics[width=1\linewidth]{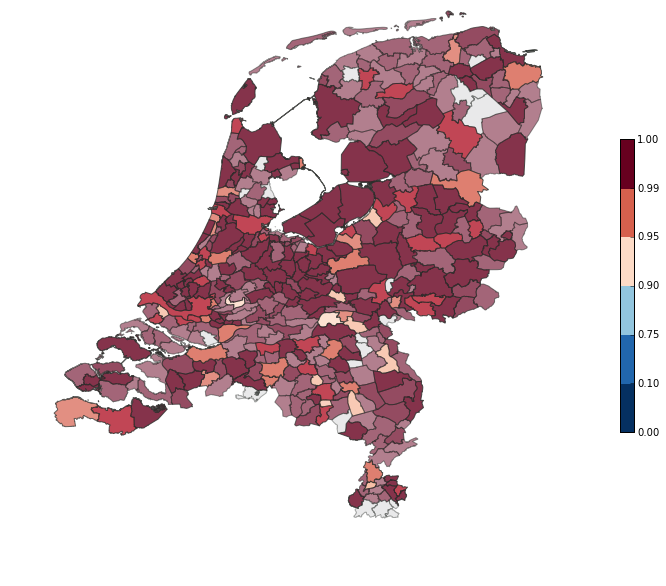}
  \caption{t-deletion (\ex{niet meer} vs. \ex{nie meer}), N=11,596}
  \label{fig:niet_meer}
\end{subfigure}

\caption{Linguistic variables on Twitter where tests disagreed}
\label{fig:twitter_sign_disagreement}
\end{figure}

Examples of linguistic variables where the approaches disagree are shown in \autoref{fig:twitter_sign_disagreement}.
The first case (\autoref{fig:doei_houdoe}) is an example of lexical variation, with two different ways of saying \ex{bye} in the Netherlands. A commonly used form is \ex{doei}, while \ex{houdoe} is known to be specific to North-Brabant, a Dutch province in the south of the Netherlands. \hsic\ and \new{join count analysis} both detect a significant pattern, but \mi\ and the Mantel test do not. The trend is less strong than in the previous examples, but the figure does suggest a higher usage of \ex{houdoe} in the south of the Netherlands. 

Another example is t-deletion for a specific phrase (\ex{niet meer} versus \ex{nie meer}), as shown in \autoref{fig:niet_meer}. Previous dialect research has found that geography is the most important external factor for t-deletion in the Netherlands, \new{with contact zones, such as the Rivers region in the Netherlands (at the intersection of the dialects of the southern province of North-Brabant, the south-west province of Zuid-Holland and the Veluwe region), having high frequencies of t-deletion~\cite{goeman1999}.}
Both \hsic\ and join count analysis report an FDR-adjusted $p<.05$, while for \mi, the geographical association does not reach the threshold of significance. 

\begin{table*}
\footnotesize
\begin{tabular}{llrlllll}
\toprule
\textbf{Linguistic variables} & \textbf{Description} & \textbf{N} & \textbf{Moran's I} & \textbf{HSIC }   & \textbf{Mantel} & \textbf{Join counts}   \\ 
\midrule
Friet / patat & french fries & 842  &  0.0004&0.0002&0.0003&0.0003\\
Proficiat / gefeliciteerd & congratulations & 14,474 & 0.0004&0.0002&0.0080&0.0003\\
Iedereen / een ieder & everyone & 13,009 & 0.8542&0.0002&0.8769&0.0432\\
Doei / aju & bye & 4,427 & 0.7163&0.0050&0.2570&0.3868\\
Efkes / eventjes&for a little while & 969 & 0.0036&0.0002&0.0003&0.0003\\
Naar huis / naar huus& to home & 3, 942 & 0.8542&0.1090&0.1245&0.9426\\
Niet meer / nie meer&not anymore & 11,596 & 0.0793&0.0002&0.5590&0.0329\\
Of niet / of nie &or not & 1,882 & 0.8357&0.1010&0.4191&0.9426\\
-oa- / -ao- &e.g.,  \emph{jao} versus \emph{joa} & 754 &  0.0004&0.0002&0.0003&0.0003\\
Even weer / weer even &for a little while again & 921 &  0.0004&0.0002&0.0003&0.0003\\
Have + participle & \specialcell[t]{e.g., \emph{heb gedaan} (`have done') \\vs. \emph{gedaan heb} (`done have')} & 1,122 &0.8587&0.2849&0.6668&0.0255\\
Be + participle& \specialcell[t]{e.g., \emph{ben geweest} (`have been') \\vs. \emph{geweest ben} (`been have')} & 1,597 & 0.0793&0.2849&0.7862&0.0051\\
Spijkerbroek / jeans& jeans &1,170 & 0.7796&0.0002&0.0080&0.0003\\
Doei/ houdoe & bye&4,491 & 0.5016&0.0002&0.6668&0.0047\\
Bellen / telefoneren & to call by telephone&4,689 &0.2730&0.0003&0.9781&0.5941\\

\bottomrule
\end{tabular} 
\caption{Twitter results. The $p$-values were calculated using 10,000 permutations and corrected for multiple comparisons.}
\label{tab:results_twitter}
\end{table*}

We also present preliminary results on using HSIC as an exploratory tool on the same Twitter corpus. To focus on active users who are most likely tweeting on a personal basis, we  exclude users with 1,000 or more followers and users who have fewer than 50 tweets, resulting in 8,333 users. We exclude infrequent words (used by fewer than 100 users) and very frequent words (used by 1000 users or more), resulting in a total of 5,183 candidate linguistic variables. We represent the usage of a word by each author as a binary variable, and use \hsic\ to compute the level of spatial dependence for each word.

\begin{figure}
\centering
\begin{subfigure}{0.4\textwidth}
  \centering
  \includegraphics[width=1\linewidth]{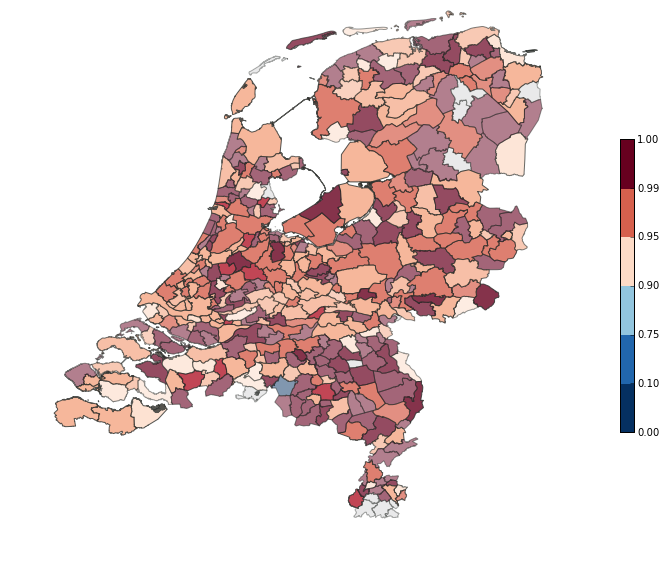}
  \caption{Usage of \emph{joh}}
  \label{fig:joh}
\end{subfigure}%
\hspace{0.2in}
\begin{subfigure}{.4\textwidth}
  \centering
  \includegraphics[width=1\linewidth]{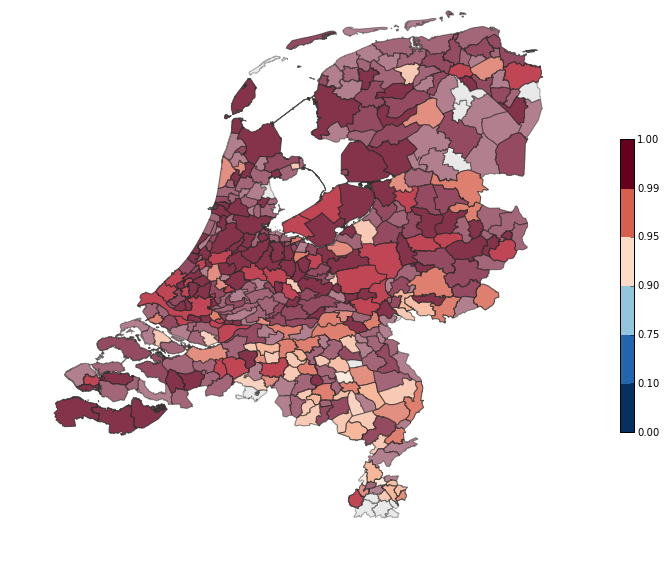}
  \caption{Usage of \ex{dadelijk}}
  \label{fig:dadelijk}
\end{subfigure}

\caption{Linguistic features on Twitter}
\label{fig:twitter_exploratory}
\end{figure}

The top 10 words with the highest \hsic\ scores are
\ex{groningen} (city), \ex{zwolle} (city), \ex{eindhoven} (city) 
\ex{arnhem} (city),  \ex{breda} (city),
\ex{enschede} (city), \ex{nijmegen} (city),
\ex{leiden} (city),
\ex{twente} (region) and \ex{delft} (city). 
While these words do not reflect dialectal variation as it is normally construed, we expect their distribution to be heavily influenced by geography. Manual inspection revealed that many English words (e.g., \ex{his, very}) have high geographical dependence. English speakers are more likely to visit tourist and commercial centers, so it is unsurprising that these words should show a strong geographical association. The top-ranked non-topical word is \ex{proficiat}, occurring at rank 34 according to \hsic. \ex{Proficiat} had previously been identified as a candidate dialect variable, and was included in our analysis in \autoref{tab:results_twitter}; this replication of prior dialectological knowledge validates the usage of \hsic\ as an exploratory tool. Less well known are \ex{joh} (an interjection) and \ex{dadelijk} (`immediately'/`just a second'), which are ranked respectively at \#60 and \#71 by \hsic. The geographical distributions of these words are shown in Figures \ref{fig:joh} and \ref{fig:dadelijk}; both seem to distinguish the southern part of the Netherlands from the rest of the country. The identification of these words speaks to the potential of \hsic\ to guide the study of dialect by revealing geographically-associated terms.\footnote{\new{The top 10 words with the highest \mi~scores are similar: \ex{groningen, eindhoven, friesland} (province), \ex{leeuwarden} (city), \ex{zwolle}, \ex{proficiat}, \ex{drachten} (city), \ex{carnaval} (a festival), \ex{brabant} (province), \ex{enschede} (city).}}

%% file: conclusion.tex
\section{Conclusion}
\label{sec:conclusion}
We have reviewed four methods for quantifying the spatial dependence of linguistic variables: \mi, which is perhaps the best-known in sociolinguistics and dialectology; \new{join count analysis}; the \mt; and the Hilbert-Schmidt Independence Criterion (\hsic), which we introduce to linguistics in this paper. \new{Of these methods, only \hsic\ is \emph{consistent}, meaning that it converges to an accurate measure of the statistical dependence between $X$ and $Y$ in the limit of sufficient data. In contrast, the other approaches are based on parametric models. When the assumptions of these models} \new{are violated, the power to detect significant geo-linguistic associations is diminished. All three of these methods can be modified to account for various geographical distributions: for example, the spatial weighting matrix employed in \mi\ and join count analysis can be constructed as a non-linear or non-monotonic function of distance~\cite{getis2010constructing}, the distances in the \mt\ can be censored at some maximum value~\cite{legendre2015should}, and so on. However, such modifications require the user to have strong prior expectations of the form of the geolinguistic dependence, and open the door to $p$-value hacking through iterative ``improvements'' to the test. In contrast, \hsic\ can be applied directly to any geotagged corpus, with minimal tuning. By representing the underlying probability distributions in a Hilbert space, \hsic\ implicitly makes a comparison across high-order moments of the distributions, thus recovering evidence of probabilistic dependence without parametric assumptions.}

\new{These theoretical advantages are borne out in an analysis of synthetic data. We consider a range of realistic scenarios, finding that the power of \mi, the \mt, and join count analysis depends on the nature of the geographical variation (e.g., dialect continua versus centers), and in some cases, even on the direction of variation. Overall, we find that \hsic, while not the most powerful test in every scenario, offers the broadest applicability and the least potential for catastrophic failure of any of the proposed approaches.} 

We then showed how to apply these tests to a diverse range of real
datasets: frequency observations in letters to the editor, binary observations in a dialect atlas, and binary observations in social media. We find that previous results on newspaper data were dependent on the procedure of selecting the geographical distance cutoff to maximize the number of positive test results; using all other test procedures, the significance of these results disappears. On the dialect atlas, we find that the fraction of statistically significant variables ranges from 55.2\% to 78.3\%  depending on the statistical approach. On the social media data, we obtain largely similar results from the four different tests, but \hsic\ detects the largest number of significant associations, identifying cases in which geography and population density were closely intertwined. 

\new{To conclude, we believe that kernel embeddings of probability measures offer a powerful new approach for corpus analysis. In this paper, we have focused on measuring geographical dependence, which can be used to test and discover new dialectal linguistic variables. But the underlying mathematical ideas may find application in other domains, such as tracking change over time~\cite{popescu2014time,vstajner2011diachronic}, or between groups of authors~\cite{koppel2002automatically}. Of particular interest for future research is the use of structured kernels, such as tree kernels~\cite{collins2001convolution} or n-gram kernels~\cite{lampos2015assessing}, which could test for structured linguistic phenomena such as variation in syntax~\cite{johannsen2015cross} or phonological change~\cite{bouchard2007probabilistic}.}


\mynewpage